\documentclass{article}

\PassOptionsToPackage{numbers, compress}{natbib}
\usepackage[final]{neurips_2025}

\usepackage[utf8]{inputenc} % allow utf-8 input
\usepackage[T1]{fontenc}    % use 8-bit T1 fonts
\usepackage{hyperref}       % hyperlinks
\usepackage{url}            % simple URL typesetting
\usepackage{booktabs}       % professional-quality tables
\usepackage{amsfonts}       % blackboard math symbols
\usepackage{nicefrac}       % compact symbols for 1/2, etc.
\usepackage{microtype}      % microtypography
\usepackage{xcolor}         % colors
\usepackage{hyperref}
\usepackage{url}
\usepackage{microtype}
\usepackage{graphicx}
\usepackage{subfigure}
\usepackage{makecell}
\usepackage{multirow}
\usepackage{colortbl}
\usepackage{xcolor}
\usepackage{booktabs} % for professional tables
\usepackage{longtable}
\usepackage{arydshln} % For dashed lines
\usepackage{xspace}
\usepackage[most]{tcolorbox}
\usepackage{wrapfig} % Add this to your preamble
\usepackage{caption}
\definecolor{mydarkblue}{rgb}{0,0.08,0.45}
\definecolor{mydarkgreen}{RGB}{0, 139, 69}
\definecolor{MAEblue}{RGB}{47 112 182}
\definecolor{SDEblue}{RGB}{28 58 88}
\hypersetup{
	colorlinks=true,
    linkcolor=blue,
	urlcolor=magenta,
	citecolor=cc4,
}
\definecolor{mycyan}{cmyk}{.3,0,0,0}
\definecolor{cc1}{rgb}{1.0, 0.44, 0.37}
\definecolor{cc2}{rgb}{0.0, 0.2, 0.6}
\definecolor{cc3}{RGB}{255, 191, 0}
\definecolor{cc4}{RGB}{0, 128, 128}
\newcommand{\ourmethod}{{SkyLadder}\xspace}

\title{SkyLadder: Better and Faster Pretraining via \\ Context Window Scheduling}

\author{
    Tongyao Zhu$^{1,2}$\;
    Qian Liu$^{2}$\thanks{Corresponding author.}\;    
    \ Haonan Wang$^{1}$\;
    Shiqi Chen$^{3}$\;\\
    \textbf{Xiangming Gu}$^{1}$\;
    \textbf{Tianyu Pang}$^{2}$\;
    \textbf{Min-Yen Kan}$^{1}$\\
    $^{1}$National University of Singapore\quad
    $^{2}$Sea AI Lab\quad
    $^{3}$City University of Hong Kong\\
    \texttt{tongyao.zhu@u.nus.edu; liuqian.sea@gmail.com}
}

\begin{document}

\definecolor{lightblue}{RGB}{173,216,230}
\definecolor{darkgreen}{rgb}{0,0.5,0}

\maketitle

\begin{abstract}
Recent advancements in LLM pretraining have featured ever-expanding context windows to process longer sequences. However, our controlled study reveals that models pretrained with shorter context windows consistently outperform their long-context counterparts under a fixed token budget. This finding motivates us to explore an optimal \textbf{context window scheduling} strategy to better balance long-context capability with pretraining efficiency. To this end, we propose \ourmethod, a simple yet effective approach that implements a short-to-long context window transition. \ourmethod preserves strong standard benchmark performance, while matching or exceeding baseline results on long-context tasks.
Through extensive experiments, we pretrain 1B-parameter models (up to 32K context) and 3B-parameter models (8K context) on 100B tokens, demonstrating that \ourmethod yields consistent gains of up to 3.7\% on common benchmarks, while achieving up to 22\% faster training speeds compared to baselines\footnote{Project code is at \url{https://github.com/sail-sg/SkyLadder}}.
\end{abstract}

\section{Introduction}\label{sec:introduction}

% There is an increasing trend to pretrain language models with longer context windows. For instance, BERT~\citep{kenton2019bert} and GPT-1 pretrained with a context window of only 512, and GPT2~\citep{radford2019language} doubled it to 1024. With the emergence of LLMs going beyond 1B parameters, the first generation of Llama1~\citep{touvron2023llama} has a context length of 2048, while the second generation increases it to 4096, and subsequently 8192 for Llama3~\citep{dubey2024llama}. It is natural that longer context window appears useful in many tasks involving long form of texts such as reading comprehension or summary. However, is pretraining with a long context window indeed a good choice for the model?

\begin{figure}
%\begin{wrapfigure}{r}{0.50\textwidth}
   % \footnotesize
%     \begin{minipage}{0.25\textwidth}
%         \centering
% \includegraphics[width=\linewidth]{figures/context-and-year.pdf}
%     \end{minipage}%
%     \begin{minipage}{0.25\textwidth}
%         \centering
%         \includegraphics[width=\linewidth]{figures/length_perf_compare_1b.pdf}
%     \end{minipage}
% \includegraphics[width=0.48\textwidth ]{figures/context-and-year-and-acc.pdf}
\centering
\includegraphics[width=0.96\textwidth ]{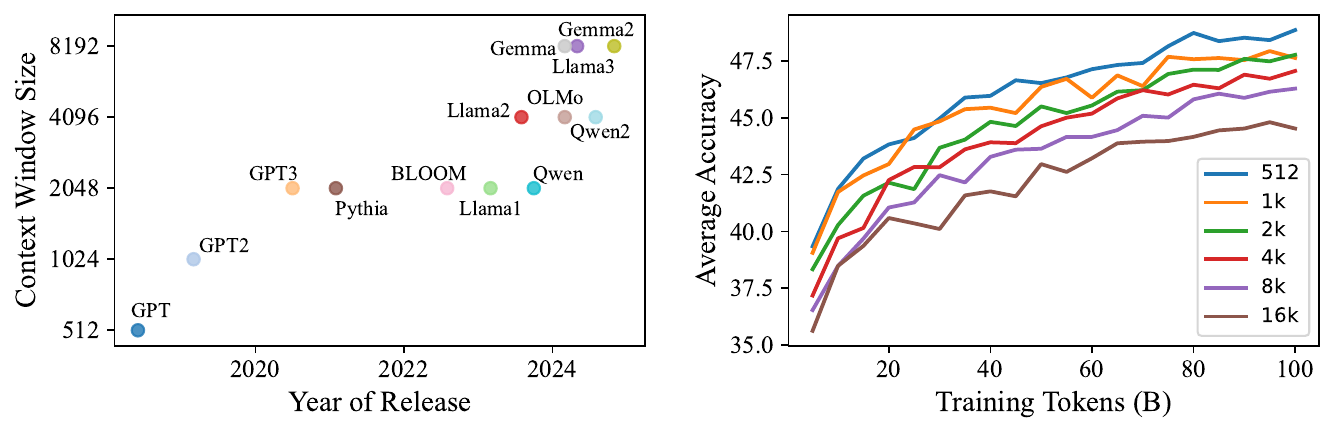}
\vspace{-2mm}
    % Min: Orig
    % \caption{Average performance across 9 popular downstream tasks for 1B-parameter models pretrained with different context window sizes. Despite increasing context window sizes over the years, longer context windows lead to degraded model performance. For example, models trained with 512-token windows and 40B tokens can achieve the similar performance as those trained with 8,192-token windows at 100B tokens.
    \caption{Left: Pretraining context window of LLMs grows over the recent years. Right: Average performance~(in \%) across nine downstream tasks for 1B-parameter models with different pretrained context window sizes (color-coded). Increasing the context window degrades the overall performance.
    % Min: moved to body text.
    % For example, models trained with 512-token windows and 40B tokens can achieve similar performance as those trained with 8,192-token windows at 100B tokens.
    % \qian{Unify the figure style of Left and Right. And not sure if we should call Phi-4 as having 16K context length since it is pretrained on 4K.}
    % \tongyao{the plots are too small.. maybe we can separate them}
    % } 
    }
    \label{fig:length_perf_compare_1b}\vspace{-4.5mm}
\end{figure}

% \qian{
The evolution of language models has been marked by a consistent expansion in context window sizes (Figure~\ref{fig:length_perf_compare_1b} left). 
While early models like GPT~\citep{radford2018improving} and BERT~\citep{kenton2019bert} were limited to context windows of 512 tokens, subsequent models have pushed significantly beyond these bounds. GPT-2~\citep{radford2019language} doubled this capacity to 1024 tokens, and with Large Language Models (LLMs) exceeding 1B parameters, this trend has continued: Llama~\citep{touvron2023llama} has a 2048-token window, followed by Llama-2~\citep{touvronllama2} (4096 tokens), and Llama-3~\citep{dubey2024llama} (8192 tokens).
% The expansion of pretraining context windows is motivated by two key factors: the requirement for models to process long sequences during inference, and the necessity to learn long-range dependencies effectively.
% \qian{The expansion is also motivated by a widespread assumption that models trained on long context windows should maintain performance parity with their short-context counterparts.}
% Original
% The push to expand pretraining context windows is primarily motivated by the growing demand for models capable of handling longer sequences during inference.
The need for models to handle longer sequences during inference has fueled the rush to expand the context window.
As models pretrained with longer context windows reduce 
document truncation and preserve % informational
coherence~\citep{ding2024fewertruncationsimprovelanguage}, there is a widespread belief that such models 
should perform comparably to, or even surpass, their shorter-context counterparts.
% Given this trend, future models will likely be pretrained on larger context windows, potentially reaching 16K or 32K tokens.
% Min: is this needed?  Omitting for now
% Given the current trajectory, future models will likely adopt pretraining context windows of 16K tokens as a default setting.

% We begin our analysis by pretraining language models with the same amount of data, but with different context window lengths. We discover a shocking phenomenon: shorter-context models generally perform better. Additionally, we consider different packing and masking strategies. Our findings reveal that under all cases, shorter context yields stronger performance. 

%\vspace{-7mm}
%\end{wrapfigure}

%%% Previous version
% In this paper, we investigate whether this trend of pretraining language models with a large context window is beneficial. We raise a fundamental question: Is pretraining with such extended context windows actually the optimal approach for developing models capable of processing long sequences? If not, what would be the best strategies to pretrain a model with a long context?

% \qian{In this paper, we challenge this trend of ...
% However, this raises a fundamental question: Is pretraining with such extended context windows actually the optimal approach for developing models capable of processing long sequences?
% }

%%% Qian's version 0128
% Min: probably we need to add back some citations here if we agree to leave out much of the first paragraph.
% Min: orig
% However, upon closer examination of previous work, we find that there is no fair experimental setup for comparing model performance across different context windows while adhering to the same token budget.
% Therefore, we first question the common belief that larger context windows do not harm performance. 
We question the common belief that larger context windows do actually improve performance. 
Close inspection of previous work reveals that there has yet to be a fair experimental setup for comparing models across different context windows while adhering to a fixed token budget.
Using tightly controlled experiments, we test how changing only the context window size during pretraining impacts their performance.
As shown in Figure~\ref{fig:length_perf_compare_1b} (right), our results indicate that models pretrained using shorter contexts always outperform long-context models, when assessed by their average performance across popular benchmarks.
% For example, models trained with 512-token windows and 40B tokens achieve comparable performance as those trained with a much larger 8,192-token window size after a 100B-token budget.
In addition, we verify that the performance gap is not eliminated by using advanced document packing strategies~\citep{dubey2024llama,ding2024fewertruncationsimprovelanguage,shicontext}.

To ensure the model can ultimately process long sequences, the model still needs to be exposed to long sequences.
However, given the finding that shorter context windows enhance performance on downstream tasks, we face a trade-off between long-context capability and pretraining effectiveness.
% Inspired by learning rate scheduling,
We propose \ourmethod, a simple yet effective \textbf{context window scheduling} strategy designed to balance both objectives. \ourmethod does this by progressively expanding the size of the context window during pretraining, beginning pretraining with a minimal short context window (e.g., 8 tokens) and progressively expanding it to the long target context window (e.g., 32,768 tokens).

% \textcolor{red}{Pending, not sure if we want to put it in. I think the analysis is a bit weak. }

% Empirical results demonstrate that \ourmethod outperforms naive long-context pretraining baselines across both short- and long-context evaluation tasks. 
% For example, models trained with our schedule achieve higher accuracy on tasks requiring localized reasoning (e.g., sentence-level classification) while maintaining strong performance on long-context benchmarks (e.g., document summarization).
% Finally, we set to analyze why the model performs better when there are short sequences inside. We find that there could be multiple explanations. First, a shorter context window means that the model has less contextual information to rely on, and therefore forces the model to memorize ngrams to predict the next token. Next, we also find that dynamically increases the context length helps the formation of 'attention sink', a phenomenon that is commonly observed across LMs of all scales.
Empirical results on 1B-parameter models (up to 32K context window) and 3B-parameter models (up to 8K context window) on 100B tokens demonstrate that \ourmethod outperforms naive long-context pretraining baselines, in both short- and long-context evaluation tasks.
For example, models trained with \ourmethod demonstrate significantly higher accuracy on standard benchmarks (e.g., HellaSwag), and reading comprehension tasks (e.g., HotpotQA), while still maintaining competitive performance on long-context evaluations like RULER.
We further investigate the mechanisms behind the superior performance by observing the training dynamics, and discover that \ourmethod exhibits more concentrated and effective attention patterns. 
Overall, we suggest that the length of the context window is an important dimension in pretraining and should be scheduled over the course of training. We recommend a progressive approach that begins with a small context of 8 tokens and gradually increases according to a linear function of training steps. Given a target context window (e.g., 32K), we suggest that allocating approximately 60\% of the total training tokens to this expansion phase leads to stronger downstream performance compared to baselines. This scheduling strategy optimally enhances both training efficiency and model capability, offering a practical recipe for improving pretraining in language models.
\vspace{-1mm}
% Overall, we hope that our study provides insights into how the choice of context window and masking strategies affect language models' pretraining. 

\section{Related Work}\label{sec:related_work}

%\begin{figure}[t]
\begin{wrapfigure}{r}{0.5\textwidth}

    \centering
    \vspace{-7.5mm}
    \includegraphics[width=1.0\linewidth]{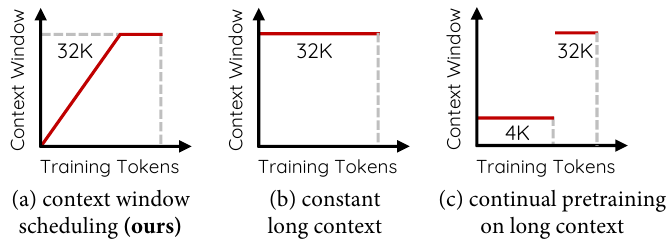}
    \caption{Schematic comparison of training-time context window scheduling.}
    \vspace{-3mm}
    \label{fig:schedule_illustration}
%\end{figure}
\end{wrapfigure}

\paragraph{Context Window Scheduling.}
Early work explored gradually increasing the context window in smaller models like BERT and GPT-2, to improve training stability and efficiency \citep{nagatsuka2021pre, li2022stabilityefficiencydilemmainvestigatingsequence, jin2023growlengthacceleratingllmspretraining}.
Notably, \citet{li2022stabilityefficiencydilemmainvestigatingsequence} proposed length warmup for more stable training but did not show clear performance gains, while \citet{jin2023growlengthacceleratingllmspretraining} focused on training acceleration in 400M models. 
We extend these findings by demonstrating, for the first time, that context window scheduling significantly boosts both \emph{efficiency} and \emph{performance} at much larger scales (up to 3B parameters).
A parallel approach from \citet{pouransari2024dataset} segments training documents by length, but \citet{fu2024dataengineeringscalinglanguage} caution that such segmentation can introduce domain biases, as longer texts often cluster in specific domains such as books. Recent developments in continual pretraining with long context windows~\citep{peng2024yarn,wang2024precision,gao2024prolong}, can also be viewed through the lens of context window scheduling with different strategies (illustrated in Figure~\ref{fig:schedule_illustration}). 
% In summary, 
Our work represents the first demonstration of both effectiveness and efficiency of context window scheduling, providing empirical evidence of its benefits in both standard and long-context benchmarks.

\paragraph{Long-Context Language Models.} Long-context language models have received a lot of attention due to their ability to capture extended dependencies across large textual windows. Most existing approaches follow a \emph{continual pretraining} paradigm~\citep{fu2024dataengineeringscalinglanguage, Xiong2023EffectiveLS}, which extends a pretrained backbone model to longer contexts through specialized fine-tuning or additional training. Several works propose to intervene in the positional embeddings to accommodate longer sequences~\citep{an2024doeseffectivecontextlength,ntk,peng2024yarn,chen2023extending,jinllm}, while others perform extended pretraining on longer-sequence corpora~\citep{gao2024prolong, wang2024precision, lu2024controlled, zhao2024longskywork}. Our approach differs from previous methods as we train \emph{native} long-context models from scratch, rather than modifying a pretrained model in post-training.
Compared with a naive long-context pretraining baseline with a constant schedule, our approach delivers substantial gains on multiple long-context tasks, underscoring the benefits of training from scratch. These findings show that our method can be a promising direction for future research on building language models with longer context windows.

\section{How Context Window Affects Pretraining}

% \sq{We first briefly introduce the background of data packing and attention masking, and then present a controlled study to examine how pretraining with varying context window sizes affects model performance. By training models with context windows ranging from 512 to 16,384 tokens under fixed computational budgets and evaluating via perplexity and downstream task benchmarks, we investigate whether longer contexts inherently improve model quality, and analyze how data packing strategies interact with context window sizes to shape conclusion.}

% In this section, 
% We now answer the question:
\emph{How does context window affect pretraining?} To investigate this in a fair and comparable manner, we pretrain language models from scratch with context windows ranging from 512 to 16,384 tokens under a fixed total number of tokens, evaluating via perplexity and downstream task benchmarks. We examine how the context window size impacts model performance, analyzing how data packing and masking strategies interact with window size.
\begin{figure*}[t]
    \centering
    \includegraphics[width=\linewidth]{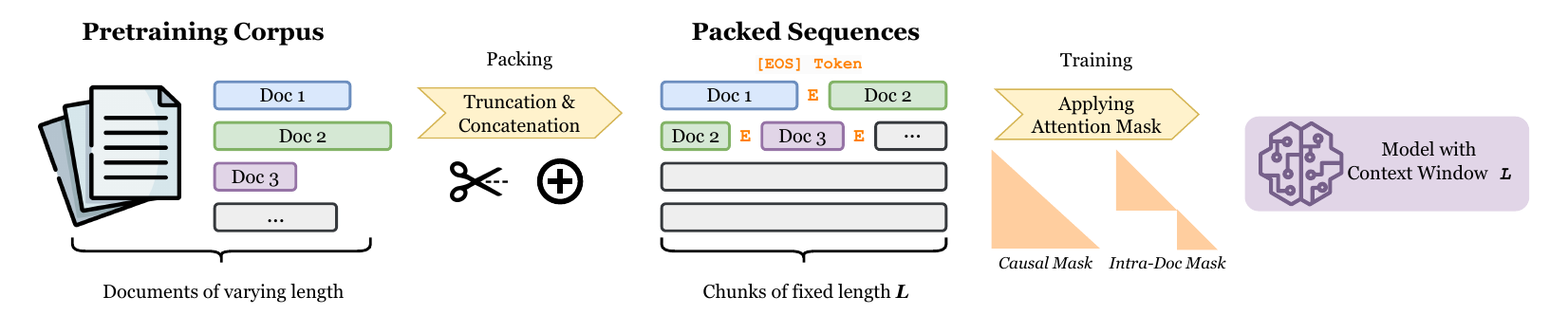} 
    \vspace{-2mm}
    \caption{An illustration of the workflow for pretraining data preparation highlights several critical decisions. Key considerations include the method of data packing, the type of attention mask to employ (causal or intra-doc mask), and determining the appropriate context window length $L$.}
    \label{fig:packing-diagram}
    \vspace{-5mm}
\end{figure*}

\subsection{Packing, Masking and Context Window}

Most modern LLMs are based on a decoder-only transformer architecture~\citep{transformer2017} with a fixed context window size denoted by $L$.
In contrast, the pretraining corpus, $D = \{d_1, d_2, d_3, \dots, d_n \}$, consists of documents with varying lengths different from $L$. Therefore, a key step before pretraining is to \textit{pack} the documents into sequences of length $L$. Formally, a packed sequence $C_i$ is constructed as $C_i = \textrm{Trunc}(d_{i,1}) \oplus d_{i,2} \oplus \dots \oplus d_{i,n-1} \oplus \textrm{Trunc}(d_{i,n})
$
% , which can be a concatenation (denoted by $\oplus$) of several documents separated by \texttt{EOS} token\footnote{We follow previous works~\citep{shicontext, zhao2024analysing} to use \texttt{EOS} tokens to denote document boundaries.}, and the $\textrm{Trunc}$ function denotes that the documents may be truncated to ensure that $\textrm{len}(C_i) = L$.
% Qian's version
, where $\oplus$ represents concatenation, and $\textrm{Trunc}(\cdot)$ denotes truncation of documents to ensure $\textrm{len}(C_i) = L$. Following previous works~\citep{shicontext, zhao2024analysing}, document boundaries within $C_i$ are explicitly marked using end-of-sequence (\texttt{[EOS]}) tokens.

% Tongyao's version
% After the sequences are packed, the inputs will be fed into transformer layers to for next-token prediction training. One of its key components is the attention mechanism, which can be written as $A_{i,j} = q_i^Tk_j$, where $\textrm{Attn}(X) = \textrm{softmax}(A+M)$. In decoder-only models, a mask $M$ will be applied to the model. A standard practice is the \textit{causal mask}, which has 
% Qian's version
After the sequences are packed, the inputs are passed into transformer layers for next-token prediction training. A crucial component of these layers is the attention mechanism, which can be formulated as 
% Min: use \top for transpose.
$A_{i,j}=q_i^\top k_j$, and then $\text{Attn}(X) = \textrm{Softmax}(A + M)$. In decoder-only models, a mask $M$ is applied to introduce constraints.
A common approach is to use a \textit{causal mask}, which ensures that each position can only attend to previous tokens by masking out (setting to $-\infty$) attention scores corresponding to future positions: $M_{ij}=-\infty$ for $j>i$ and $M_{ij}=0$ otherwise.
% \[
% M_{i,j} = 
% \begin{cases} 
% 0 & \text{if } i \geq j  \\
% -\infty & \text{otherwise}
% \end{cases}
% \]
A recently proposed masking scheme, known as \textit{intra-doc} mask~\citep{zhao2024analysing,dubey2024llama}, imposes a constraint that only allows tokens to attend to each other if they belong to the same document. Let each document $d$ have start index $s_d$ and end index $e_d$, the masking can be denoted as $M^{\text{intra}}_{ij}=0$ when $\exists\,d$ such that $s_d\le i,j\le e_d$ and $j\le i$, and $M^{\text{intra}}_{ij}=-\infty$ otherwise. The model is trained with the standard cross-entropy loss on the packed sequences of length $L$. The workflow for pretraining data processing is illustrated in Figure~\ref{fig:packing-diagram}.

\subsection{Preliminary Study on Context Window Size}\label{subsec:exp-setup}

% Our preliminary study reveals that BM25 gives higher relevance than Contriever, as shown in Appendix X. Therefore, we choose BM25 to represent semantic packing. 

% As introduced in Section~\ref{sec:introduction}, we begin our study from investigating the impact of context window size on model performance by conducting a controlled study.
% Specifically, we pretrain language models with varying context window sizes while keeping all other experimental settings unchanged.
% This approach allows us to isolate the effects of context window size and systematically analyze its influence on model performance.
% Through this analysis, we aim to determine whether longer context windows inherently lead to improved performance or if other factors come into play.

%%%% Qian's version
As per Section~\ref{sec:introduction}, we initiate our study by investigating the impact of context window size on model performance through a controlled experiment.
Specifically, we pretrain language models with varying context window sizes, while preserving all other experimental settings.
This enables a pure analysis of the context window's influence on model performance.
Through this analysis, we aim to understand whether \textit{longer context windows inherently lead to better or worse model performance}.

\paragraph{Key Variables.}
The context window size determines the number of tokens included in the context for each packed sequence.
However, as discussed earlier, several additional factors influence the content within the context window: (1) \textit{Packing methods} determine which documents constitute the context window, and different packing strategies can significantly alter the composition of token sequences; (2) \textit{Masking methods} decide whether cross-document attention is enabled within the same context window. The choice of masking affects how the information from different documents interacts during training.\looseness=-1

% \paragraph{Packing} To study the impact of packing, we employ two widely adopted strategies: random packing and semantic packing. For random packing, there is no particular ordering of the documents. For semantic packing, we reproduce the procedure used by~\citet{shicontext} to retrieve and concatenate relevant documents from the corpus. We experimented with both a dense retriever, Contriever~\citep{izacard2021contriever}, as discussed in the original paper, as well as a classical sparse retriever, BM25. 

\paragraph{Packing and Masking.} To study the impact of packing, we employ two strategies: \textit{random packing} and \textit{semantic packing}. 
For random packing, documents are randomly concatenated without a specific ordering. 
For semantic packing, inspired by \citet{shicontext}, we retrieve and concatenate semantically relevant documents from the corpus, aiming to keep them within the same context window.
% After experimenting with both a dense retriever, Contriever~\citep{izacard2021contriever} and a typical sparse retriever BM25, we find that BM25 is much more robust and thus we primarily focus on BM25 for our experiments here.
After experimenting with both a dense retriever~\citep{izacard2021contriever}  and a lexical retriever BM25, we found that BM25 gives stronger performance and chose it as our focus.
For masking, the baseline approach is causal masking, where each token can attend to all preceding tokens within the same context window, regardless of document boundaries.
Conversely, recent studies~\citep{zhao2024analysing,ding2024fewertruncationsimprovelanguage} show that disabling cross-document attention, thereby enabling intra-document attention, improves performance.
For clarity in subsequent discussions, we denote random packing with causal masking as \textbf{Random}, BM25 packing with causal masking as \textbf{BM25}, and random packing with intra-document masking as \textbf{IntraDoc}.

\label{para:training_setup}
\paragraph{Training.} We pretrain models from scratch using the TinyLlama codebase~\citep{zhang2024tinyllamaopensourcesmalllanguage}, and study models with 120M, 360M and 1B parameters.
Given the substantial computational cost associated with retrieval in semantic packing, we randomly select around 30B tokens from the CommonCrawl (CC) subset of the SlimPajama dataset~\citep{cerebras2023slimpajama} as the pretraining corpus. All models undergo training for up to 100B tokens ($\sim$3.3 epochs).
To ensure consistency across experiments, we strictly control all other settings, retaining the same batch size and learning rate schedule for all context windows.
All models also incorporate Rotary Positional Encoding (RoPE)~\citep{su2024roformer} to encode positional information. 
Appendix~\ref{app:sec:model-architecture} and~\ref{app:sec:training-config} give further model architecture details and training settings.
\looseness=-1
% \paragraph{Evaluation} As we are pretraining language models from scratch, we consider perplexity on validation documents as a key metric following prior studies to compare language model pretraining. A loss difference larger than 0.01 is considered significant~\citep{fu2024dataengineeringscalinglanguage}. 
% When we compare two models of different lengths (e.g. a 2k-context model and an 8k-context model), we ensure that the evaluation sequence is within the context length for the shorter model, to avoid the need to extend a short-context model. For downstream tasks, we additionally test the performance on common benchmarks, including hellaswag~\citep{zellers2019hellaswag}, ARC-easy~\citep{allenai:arc} and challenge, winogrande~\citep{sakaguchi2021winogrande}, CommonsenseQA~\citep{talmor-etal-2019-commonsenseqa}, OpenbookQA~\citep{OpenBookQA2018}, PIQA~\citep{PIQA}, Social-IQA~\citep{sap-etal-2019-social}, and MMLU~\citep{mmlu}. We use the OLMES~\citep{OLMES} suite for measuring performance, as~\citet{gao2024prolong} show that it offers high reliability and stability with manually curated 5-shot demonstrations. 

\paragraph{Evaluation.} For all model sizes, we use perplexity (PPL) on validation documents from the original dataset as a key metric, in line with established practices~\citep{fu2024dataengineeringscalinglanguage,kaplan2020scaling,hoffmann2022training}.
% A loss difference greater than 0.01 can be considered significant \citep{fu2024dataengineeringscalinglanguage}.
Note that when comparing models across different context windows (e.g., a 2K-context model and an 8K-context model), we must ensure the evaluation sequence fits within the shorter model's context window to maintain a fair comparison.
We also evaluate 1B models on downstream standard benchmarks: HellaSwag \citep{zellers2019hellaswag}, ARC-Easy and ARC-Challenge \citep{allenai:arc}, Winogrande \citep{sakaguchi2021winogrande}, CommonsenseQA \citep{talmor-etal-2019-commonsenseqa}, OpenBookQA \citep{OpenBookQA2018}, PIQA \citep{PIQA}, Social-QA \citep{sap-etal-2019-social}, and MMLU \citep{mmlu}. We employ the OLMES suite \citep{OLMES} for the evaluation, as it has been shown to provide reliable and stable results with curated 5-shot demonstrations \citep{gao2024prolong}.\looseness=-1

% \subsection{Results}

% We show the results. 

% \paragraph{Intra-document masking gives the best performance.} As shown in , intra-document attention outperforms full-attention by a large margin, which is consistent with previous findings by~\citet{zhao2024analysing}. In addition, we also observe that intra-document attention (Random-Intra) is also better than relevance-based concatenation with a fully-open mask (BM25-Full). This is surprising as BM25 requires much more compute that random packing to perform global retrieval. Therefore, from a practical perspective, Intra-Document masking is an efficient yet effective approach to improve performance. 

% However, an often-neglected factor is that the intra-document masking, in general, provides much shorter sequences. In Figure xxx, we show the actual length of the data distribution. Overall, the real context length of intra-document masking models is significantly shorter than the real ones. We therefore wonder if the context length $L$ itself also plays a role in the superior performance of the IntraDoc model. 

% \section{Context Length in Pretraining}

\subsection{Experimental Results}
\begin{figure*}[t]
    \centering
    % \includegraphics[width=0.32\textwidth]{figures/intradoc-length-comp-1b.pdf}
    % \hspace{0.01\textwidth}
    % \includegraphics[width=0.32\textwidth]{figures/bm25-length-comp-1b.pdf}
    % \hspace{0.01\textwidth}
\includegraphics[width=1\textwidth]{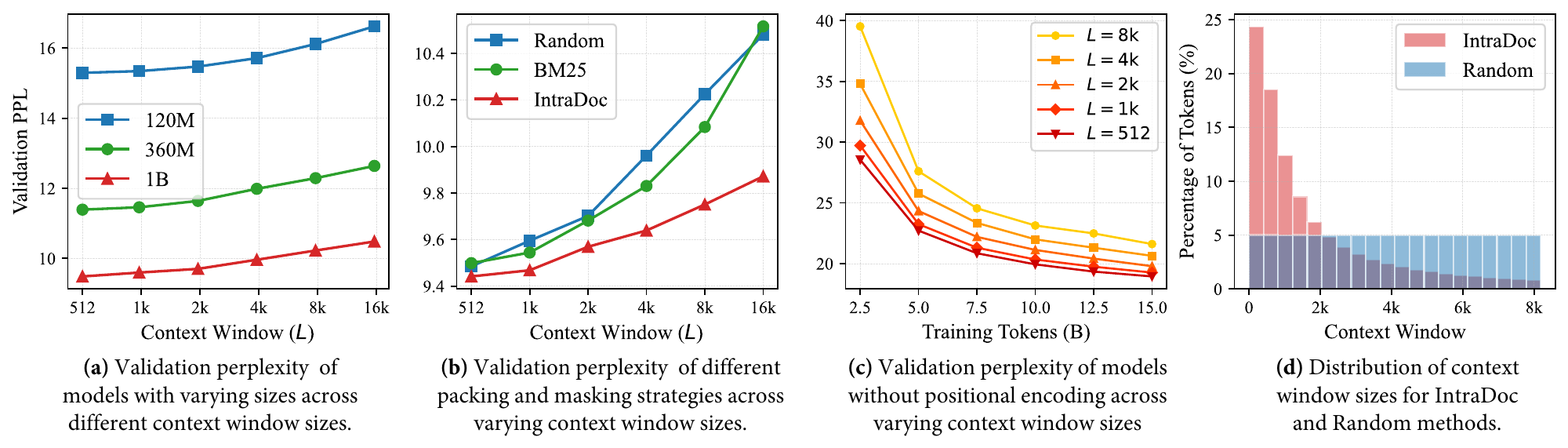}
\vspace{-3.5mm}
    % \caption{\textbf{Left}: validation PPL of models of different model sizes and context window sizes. \textbf{Right}: validation PPL of models trained with different packing and masking strategies, as well as context window sizes. Evaluations are done on the validation documents with the sliding window of 512.}
    \caption{Ablation studies of different factors on different context window sizes. Note that the validation PPL is obtained on the validation documents with a sliding window size of 512 tokens. The packing strategy in (a) is Random, and the model sizes in (b) and (c) are 1B and 120M, respectively. Note that the context window in (d) means the number of available preceding tokens when making next-token prediction (calculation details in Section~\ref{app:sec:definition-of-context}).}
    \label{fig:packing-length-compare}\vspace{-5mm}
\end{figure*}

% \begin{figure*}[ht]
%     \centering
%     \includegraphics[width=0.38\textwidth]{figures/length-compare-ppl-1b.pdf}
%     \hspace{0.01\textwidth}
%     \includegraphics[width=0.38\textwidth]{figures/length_perf_compare_1b.pdf}
%     \hspace{0.01\textwidth}
%     % \includegraphics[width=0.32\textwidth]{figures/size-compare-ppl-v2.pdf}
%     \caption{Left: Validation perplexity of 1B models (evaluated on a sliding window of length 512) trained with different context lengths. Middle: Average downstream task performance of 1B models with different context lengths. Right: Comparison of models of different sizes and context lengths.}
%     \label{fig:length-compare-1b}
%     % Min: Use different color scheme and line quality to denote a clear series of values.  Check with B/W copies.
% \end{figure*}

% Our experimental results are presented in Figure~\ref{fig:length-compare-1b} and~\ref{fig:packing-length-compare}. From these results, we can draw the following conclusions.

Figure~\ref{fig:length_perf_compare_1b} presents the main experimental result, obtained using the Random setting with 1B-parameter models.
The results indicate that context window size significantly influences the performance of LLMs, with \textit{shorter contexts generally leading to better performance}. To further investigate the factors contributing to the observation, we perform a comprehensive analysis to examine potential variables that may affect the conclusion.
Figure~\ref{fig:packing-length-compare} shows our results, and we derive four key findings: 
\begin{tcolorbox}[colback=lightblue!10]
\textbf{Findings:}
(1) The advantage of training on shorter contexts is consistent across model sizes; 
(2) This advantage is independent of the packing and masking methods employed; 
(3) It is also unrelated to the use of positional encoding; 
(4) The best packing and masking strategy is IntraDoc, which outperforms others probably because it introduces a larger number of short contexts during pretraining. 
\end{tcolorbox}
\vspace{-0.2mm}

\paragraph{Findings (1) and (2).}
As shown in Figure~\ref{fig:packing-length-compare}, regardless of the model size in (a) or the packing and masking methods in (b), a shorter context window for pretraining generally results in higher average performance on benchmarks.
The finding on benchmarks is consistent with the trend of validation PPL, where shorter context windows always yield lower PPL.

\paragraph{Finding (3).}
When using shorter context windows, one might hypothesize that the model learns positional encoding patterns for nearer positions more frequently, leading to better performance on standard benchmarks.
To test the hypothesis, we systematically ablate RoPE by completely excluding it during pretraining, following prior work~\citep{kazemnejad2023impactpositionalencodinglength}.
In Figure~\ref{fig:packing-length-compare}(c), models trained with short-context windows still outperform their long-context counterparts, even in the absence of positional encoding. This suggests that the advantages of shorter contexts are independent of positional encoding.

\paragraph{Finding (4).}
% From Figure~\ref{fig:packing-length-compare} (b), we observe that IntraDoc achieves the best validation PPL across all context window sizes compared to Random and BM25, and we also find a consistently higher standard benchmark performance (further details on benchmark performance can be found in Appendix \qian{Tongyao: move it to appendix and echo here}).
From Figure~\ref{fig:packing-length-compare}(b), we observe that IntraDoc achieves the best validation PPL across all context window sizes compared to Random and BM25, alongside consistently higher performance on standard benchmarks (c.f. Appendix~\ref{app:context-length-study}). This raises the question: why does IntraDoc excel?
We attribute the advantage to the context window size distribution of IntraDoc, which implicitly increases the prevalence of shorter contexts. As illustrated in Figure~\ref{fig:packing-length-compare}(d), despite the sequence length of 8K, fewer than 1\% of context windows actually reach this limit.
While prior work links the success of IntraDoc to reduced contextual noise~\citep{zhao2024analysing}, we identify a complementary factor --- reduced average context window size --- as a key factor in its strong performance.
That is, we hypothesize that the effectiveness of IntraDoc may also be closely tied to short context windows.\vspace{-1mm}
\section{\ourmethod: Context Window Scheduling}

We now present~\ourmethod for progressively expanding the context window during pretraining.

\subsection{Method}
\begin{wrapfigure}{R}{0.5\textwidth}

    \centering
    \vspace{-12mm}
    \includegraphics[width=0.5\textwidth]{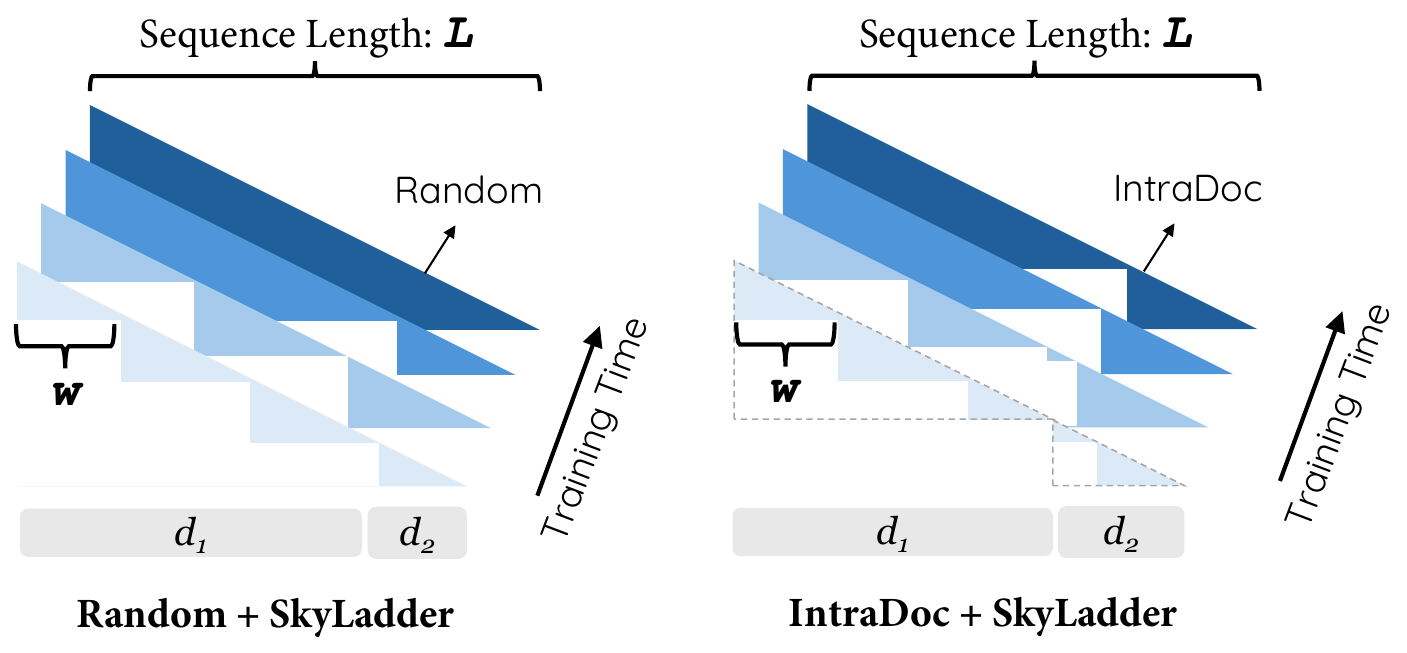} % Replace with your image file name and extension
    % \caption{An illustration of different masking strategies. The example shows a packed sequence (length $L$) consisting of two documents. For Dynamic masking, the $w$ can vary across training steps, and it can be combined with intra-document masking to preserve document boundaries. \tongyao{pending,change method name in the figure} }
    \caption{An illustration of \ourmethod with Random and IntraDoc. The example shows a packed sequence (length $L$) consisting of two documents. For \ourmethod, the context window $w$ starts from a small value and dynamically adjusts during training, eventually converging to the masking patterns of Random or IntraDoc.}
    \label{fig:masking-strategy}
    \vspace{-6mm}
%\end{figure}

\end{wrapfigure}

Inspired by learning rate scheduling, we explore whether dynamically scheduling the context window from short to long during pretraining could lead to performance improvements.
This method can be implemented by applying multiple local ``mini'' causal masks to a long, packed sequence.
We illustrate this masking strategy in Figure~\ref{fig:masking-strategy}.

Formally, we define a local window length $w$. The associated mask $M_w$ is defined as follows:
$M_{ij}=0$ when $\lfloor \frac{i}{w} \rfloor\,w \le j \le i$, and $M_{ij}=-\infty$ otherwise,
% \small
% $
% M_{i,j} = 
% \begin{cases} 
% 0 & \text{if } \lfloor \frac{i}{w}\rfloor w \leq j \leq i \\
% % 0 & \text{if } 0 \leq i-j \leq i\bmod w \\
% -\infty & \text{otherwise.}
% \end{cases}
% $\normalsize
where $\lfloor \frac{i}{w}\rfloor w$ calculates the largest multiple of $w$ that is less than or equal to $i$, effectively defining a block-wise attention mask for the query token at position $i$. We linearly adjust the window size upwards by a constant factor per training step $t$: $w(t) = \min(w_e, w_s + \lfloor \alpha t \rfloor)$, where $w_e$ and $w_s$ represent the ending and starting context window sizes, respectively. Here, $\alpha$ denotes the rate of expansion, and $t$ corresponds to the training step. As the training progresses, when the dynamic context window size $w(t)$ eventually reaches the desired (long) context window size $L = w_e$, it remains fixed at that value. At this point, the attention mask is equivalent to a full causal mask. Notably, this method modifies the effective context window through masking, independent of how the sequences are packed.
As such, this mask $M_w$ can be integrated with $M^{\textrm{Intra}}$, which maintains the attention boundaries between documents; it can be seamlessly combined with most packing and masking strategies.
% The $w_e$ should be set to the final context window size $L$, to achieve the final training goal of having a $L$ context model.
% When $w(t)$ reaches $L$, it stays there and the mask is equivalent to a full causal mask.
% To achieve the ultimate training objective of obtaining a model with a context window size of $L$, the value of $w_e$ should be set equal to $L$. 
% As the training progresses, when the dynamic context window size $w(t)$ eventually reaches $L$, it remains fixed at that value. At this point, the attention mask effectively becomes equivalent to a full causal mask.

% The mask $M_w$ can be combined with $M^{\textrm{Intra}}$, which preserves the boundary of attention between documents. It is worth noting that the method changes the actual context length by masking, which is agnostic to how the packed sequences look like: it can be combined with any possible packing method, such as BM25-based packing introduced earlier.\looseness=-1

\subsection{Experimental Setup}
\label{sec:skyladder-exp-setup}
We follow the same setup in Section~\ref{subsec:exp-setup} to pretrain language models with 8K context on 100B tokens. We set $w_s = 32$ and $\alpha = 1/8$ by default, which means that a model roughly needs 64K steps (around 64B tokens) to reach the final desired context window of $L=8192$. All baseline and \ourmethod models are implemented with Flash Attention 2~\citep{dao2023flashattention2} (pseudocode in~\ref{app:sec:implementation}).
We fix all other hyperparameters, such as the learning rate schedule, batch size, etc., for fair comparison. Due to resource constraints, we do not perform extensive hyperparameter search to obtain the best combinations for $w(t)$, $\alpha$, and $w_s$. In our ablation study, we show that these hyperparameters have a negligible impact on performance, as long as they are within a reasonable range.

For evaluation, we use the same suite mentioned in Section~\ref{subsec:exp-setup} with standard benchmarks.
To evaluate the performance of long-context question answering within an 8K 
 length, we utilize the 30-document setting from the Multi-Document QA (MDQA) benchmark \citep{liu-etal:2023:arxiv}. This is a widely-adopted benchmark that is shown to be reliable for models of 1B scale~\citep{pouransari2024dataset, zhao2024analysing}, with an average length of approximately 6K tokens.
We also select synthetic tasks within RULER~\citep{hsieh2024ruler}, as defined by~\citet{yen2024helmetevaluatelongcontextlanguage}. We choose the setup of the task that fills up the model's target context window $L$. \vspace{-1mm}
% The questions are from the NaturalQuestions-Open dataset~\citep{kwiatkowski-etal-2019-natural, lee-etal-2019-latent}. The longest task of 30-document QA has around 6k tokens on average.
% \qian{@Tongyao: add the evaluation benchmarks of helmet ruler.}

\subsection{Experimental Results}

% \begin{table*}[bt]
% \centering
% \small
% \caption{Performance (accuracy in \%) of different 1B models pretrained on 100B CC tokens on standard benchmarks. \_ denotes statistical significance from the baseline (described in \S\ref{app:sec:sky-evaluation}).}
% \setlength{\tabcolsep}{3.5pt}
% {
% \begin{tabular}{l l c c c c c c c c c}\toprule
%   \textbf{Method} 
%     & {\textbf{Avg.}} 
%     & { \textbf{ARC-E}} 
%     & { \textbf{ARC-C}} 
%     & { \textbf{CSQA}} 
%     & { \textbf{HS}} 
%     & { \textbf{OBQA}} 
%     & { \textbf{PIQA}} 
%     & { \textbf{SIQA}} 
%     & { \textbf{WG}} 
%     & { \textbf{MMLU}} \\ \midrule
% Random            & 46.3 & 58.0 & 32.7 & 49.6 & 43.0 & 40.2 & 64.8 & 46.4 & 51.9 & 29.9 \\
% +~\ourmethod     & \textbf{50.0} {\scriptsize (+3.7)} & \textbf{\underline{65.4}}
% & \textbf{\underline{35.6}}
% & \textbf{\underline{56.8}}
% & \textbf{\underline{47.0}}
% & \textbf{42.8}
% & 64.8
% & \textbf{\underline{48.9}}
% & {\underline{56.0}}
% & \textbf{\underline{32.4}} \\
% IntraDoc          & 47.4 & 61.8 & 33.4 & 52.7 & 45.6 & 38.0 & 64.3 & 45.7 & 54.8 & 30.5 \\
% +~\ourmethod     & 49.3 {\scriptsize (+1.9)} & \underline{64.8}
% & 33.8
% & \underline{55.4}
% & \textbf{\underline{47.9}}
% & 39.4
% & \textbf{\underline{66.1}}
% & \underline{48.0}
% & \textbf{56.4}
% & \underline{31.8} \\
% \bottomrule
% \end{tabular}
% }
% \label{tab:standard-tasks-1b-cc-model-performance}
% \vspace{-1mm}
% \end{table*}

\begin{table*}[bt]
\centering
\small
\caption{Performance (accuracy in \%) of different 1B models pretrained on 100B CC tokens on standard benchmarks. $^*$ denotes statistical improvements over the baseline (described in \S\ref{app:sec:sky-evaluation}).}
\setlength{\tabcolsep}{3.5pt}
{
\begin{tabular}{l l l l l l l l l l l}\toprule
  \textbf{Method}
    & {\textbf{Avg.}}
    & { \textbf{ARC-E}}
    & { \textbf{ARC-C}}
    & { \textbf{CSQA}}
    & { \textbf{HS}}
    & { \textbf{OBQA}}
    & { \textbf{PIQA}}
    & { \textbf{SIQA}}
    & { \textbf{WG}}
    & { \textbf{MMLU}} \\ \midrule
Random            & 46.3 & 58.0 & 32.7 & 49.6 & 43.0 & 40.2 & 64.8 & 46.4 & 51.9 & 29.9 \\
+~\ourmethod     & \textbf{50.0} {\scriptsize (+3.7)} & \textbf{65.4}$^*$
& \textbf{35.6}$^*$
& \textbf{56.8}$^*$
& \textbf{47.0}$^*$
& \textbf{42.8}
& 64.8
& \textbf{48.9}$^*$
& 56.0$^*$
& \textbf{32.4}$^*$ \\
IntraDoc          & 47.4 & 61.8 & 33.4 & 52.7 & 45.6 & 38.0 & 64.3 & 45.7 & 54.8 & 30.5 \\
+~\ourmethod     & 49.3 {\scriptsize (+1.9)} & 64.8$^*$
& 33.8
& 55.4$^*$
& \textbf{47.9}$^*$
& 39.4
& \textbf{66.1}$^*$
& 48.0$^*$
& \textbf{56.4}
& 31.8$^*$ \\
\bottomrule
\end{tabular}
}
\label{tab:standard-tasks-1b-cc-model-performance}
\vspace{-2mm}
\end{table*}

\begin{table*}[bt]
\centering
\small
\caption{Performance (accuracy in \%) of 1B models pretrained on 100B CC tokens with different methods on reading comprehension and long‐context benchmarks. Detailed setup is in Appendix~\ref{app:sec:sky-evaluation}.}
\begin{tabular}{l l c c c c c c c c}
\toprule
\multirow{2}{*}{\textbf{Method}}
  & \multicolumn{6}{c}{\textbf{Reading Comprehension Benchmarks}}
  & \multicolumn{3}{c}{\textbf{Long Benchmarks}} \\
\cmidrule(lr){2-7} \cmidrule(lr){8-10}
  & {\textbf{Avg.}} & {\scriptsize \textbf{HotpotQA}} 
  & {\scriptsize \textbf{SQuAD}} & {\scriptsize \textbf{NQ}} 
  & {\scriptsize \textbf{TriviaQA}} & {\scriptsize \textbf{RACE-h}}
  & {\textbf{Avg.}} & {\scriptsize \textbf{MDQA}} & {\scriptsize \textbf{RULER}} \\
\midrule
Random 
  & 25.5 & 6.5  & 37.0 & 15.8 & 37.7 & 30.7 
  & \textbf{15.3} & 17.7 & \textbf{12.8} \\
+~\ourmethod 
  & \textbf{30.2} {\scriptsize (+4.7)} & \textbf{12.4} 
  & \textbf{40.2} & \textbf{20.4} & \textbf{43.0} & \textbf{35.0} 
  & 14.3 & \textbf{18.3} & 10.3 \\
IntraDoc 
  & 28.7 & 11.4 & 39.0 & 18.2 & 42.3 & 32.3 
  & 13.0 & 15.3 & 10.6 \\
+~\ourmethod 
  & 29.1 {\scriptsize (+0.4)} & 11.0 
  & 38.5 & \textbf{20.4} & 41.5 & 34.3 
  & 13.2 & 15.6 & 10.7 \\
\bottomrule
\end{tabular}
\label{tab:1b-cc-reading-rag-long}
\vspace{-4mm}
\end{table*}

Tables~\ref{tab:standard-tasks-1b-cc-model-performance} and~\ref{tab:1b-cc-reading-rag-long} present the main results, highlighting significant improvements achieved by \ourmethod across standard benchmarks, reading comprehension tasks and long-context benchmarks. For instance, compared to the Random baseline, integrating \ourmethod yields notable performance gains on standard tasks such as MMLU (+2.5\%), ARC-E (+7.4\%), and HellaSwag (+4\%). This suggests that models with \ourmethod excel at learning common knowledge during pretraining. Additionally, our method further improves the performance of the strong baseline IntraDoc across many benchmarks. Meanwhile, for realistic long-context benchmarks like MDQA, \ourmethod matches or exceeds baseline performance. For RULER, the performance difference is likely because of fluctuation caused by its synthetic nature and small size~\citep{wang2024precision}. More long-context evaluation can be found in Section~\ref{app:sec:sky-evaluation}, confirming that \ourmethod is comparable with or better than baselines on long-context evaluation. In addition to Random and IntraDoc, we also verify that \ourmethod improves the performance of the BM25 model on both short and long tasks (Section~\ref{app:sec:additional-ablation}).

To address potential concerns that the benefits observed in short contexts may stem from the high level of noise in CC, we conduct additional experiments using the FineWeb-Pro dataset~\citep{zhou2024programming}, a carefully curated high-quality dataset containing 100B tokens. As shown in Table \ref{tab:1b-finewebpro}, improved data quality indeed leads to substantial performance gains. However, our key findings remain consistent: IntraDoc continues to outperform Random, and \ourmethod consistently delivers significant improvements over both baselines. This demonstrates that our method generalizes to corpora of varying quality.

We further examine whether \ourmethod is generalizable beyond natural language tasks. Following~\citet{ding2024fewertruncationsimprovelanguage}, we pretrain 1B code models on 100B Python code with the Starcoder tokenizer~\citep{li2023starcoder}. We observe a lower training loss ($\sim0.9$) for code pretraining compared to natural language ($\sim2.1$), suggesting that the structure in code makes the training easier. However, as shown in Table ~\ref{tab:coding-results}, there is still significant improvement when applying \ourmethod under both greedy decoding and sampling setups, especially when the target context length is 32K. This demonstrates the potential of \ourmethod to coding and possibly other reasoning tasks beyond natural language modelling. 

\begin{table*}[bt]\centering
\small

\caption{Performance (in \%) of 1B models pretrained on 100B Python code data. We follow the protocol of \citet{huang2024opencoderopencookbooktoptier} to evaluate on HumanEval~\citep{chen2021codex} and BigCodeBench~\citep{zhuo2024bigcodebench}. $t$ is the sampling temperature. \ourmethod shows consistent improvement especially for 32K-context models.}
\begin{tabular}{llrrrrrrr}\toprule
& &\multicolumn{3}{c}{\textbf{HumanEval}} &\multicolumn{3}{c}{\textbf{BigCodeBench}} \\ \cmidrule(lr){3-5}\cmidrule(lr){6-8}
& &Greedy &\multicolumn{2}{c}{Sampling ($t=0.8$)} &Greedy &\multicolumn{2}{c}{Sampling ($t=0.8$)} \\\cmidrule(lr){3-5}\cmidrule(lr){6-8}
$L$& Method &Pass@1 &Pass@10 &Pass@100 &Pass@1 &Pass@10 &Pass@20 \\\midrule
32K &Random &17.7 &32.4 &51.8 &9.0 &16.1 &19.7 \\
&+~\ourmethod &\textbf{21.3} &\textbf{37.7} &\textbf{59.8} &\textbf{9.4} &\textbf{20.6} &\textbf{24.3} \\
8K &Random &22.0 &37.2 &61.0 &9.9 &19.3 &23.6 \\
&+~\ourmethod &\textbf{23.2} &\textbf{38.2} &\textbf{63.4} &\textbf{11.3} &\textbf{20.0} &\textbf{24.1} \\
\bottomrule
\end{tabular}
\label{tab:coding-results}
\end{table*}

\subsection{Scalability Experiments}
% \paragraph{Data Quality} One might argue that the benefits of short contexts arise from the high level of noise in the CommonCrawl corpus. To address this concern, we use the Fineweb-Pro dataset~\citep{zhou2024programming}, a heavily cleaned and carefully curated high-quality dataset containing 100B tokens. As shown in Table~\ref{tab:1b-model-olmes}, the improved data quality leads to substantial performance gains across most benchmarks. However, our key conclusions remain unchanged: IntraDoc attention outperforms fully open attention, and dynamic masking continues to yield significant improvements. This demonstrates that our method generalizes well across different types of corpora, regardless of noise levels.
% In this section, we explore whether~\ourmethod can scale from two perspectives: model size, and context window size.

% In this section, we investigate the scalability of \ourmethod from two dimensions: model size and context window size.
% Specifically, we examine whether the performance improvements achieved by \ourmethod persist as we scale up the model parameters and extend the context window size. \looseness=-1
We examine whether \ourmethod's improvements persist as we scale up the model parameters and extend the context window size. We use the largest model and context size that our compute permits.

% \paragraph{Model Size} To study whether~\ourmethod could extend to larger models, we additionally train a 3B model (details in Appendix ~\shiqi{add ref}) on the high-quality FineWeb-Pro dataset over 100B tokens~\citep{zhou2024programming}. We show the results in Table~\ref{tab:model-size-scaling}. We can see that models with dynamic masking perform the best across almost all tasks. Besides, we also verify that on smaller models of size 120M and 360M, our method also leads to consitently better performance compared with the baselines. 

\paragraph{Model Size.} 
% To evaluate the scalability of \ourmethod to larger models, 
We conduct experiments across three model sizes: 120M, 360M, and 3B parameters on the Fineweb-Pro dataset. Table \ref{tab:model_size_scaling} demonstrates that models utilizing \ourmethod consistently achieve better standard benchmark performance on all model sizes. For long context tasks, our method does not benefit 120M models, possibly due to their limited capacity in processing long sequences. However, the performance gain on 3B models is prominent. We observe a positive scaling trend: as the model size grows, the performance improvement also increases, indicating the potential of applying our method to even larger models beyond our current scale. We leave it as a future work to explore larger models as it requires significantly more compute. 

\begin{table*}[tb]
\centering
\small
\begin{minipage}[t]{0.44\linewidth}
\caption{Performance (average accuracy over tasks, in \%) of 1B models pretrained on FineWeb-Pro with an 8K context window.}\label{tab:1b-finewebpro}
\begin{tabular}{lll}
\toprule
\textbf{Method} &\textbf{Standard} & \textbf{Long } \\\midrule
Random & 52.5 & 11.1 \\
~+~\ourmethod & {\textbf{55.2}} {\scriptsize (+2.7)} &{12.3} {\scriptsize (+1.2)}\\
IntraDoc & 54.3 & 12.7 \\
~+~\ourmethod &{54.8} {\scriptsize (+0.5)} &{\textbf{13.9}} {\scriptsize (+1.2)}\\
\bottomrule
\end{tabular}
\end{minipage}\quad\quad
\begin{minipage}[t]{0.5\textwidth}
\small
\caption{Performance (average accuracy in \%) for models of different sizes.}\label{tab:model_size_scaling}

    \begin{tabular}{llll}
\toprule
\textbf{Size} & \textbf{Method} &\textbf{Standard} & \textbf{Long} \\\midrule
\multirow{2}{*}{120M} &Random &40.1 &5.8 \\
&~+~\ourmethod &{41.2} {\scriptsize (+1.1)} &{5.1} {\scriptsize (-0.7)}\\
\multirow{2}{*}{360M} &Random & 47.2 & 8.9 \\
&~+~\ourmethod &{49.6} {\scriptsize (+2.4)} &{8.9} \\
\multirow{2}{*}{3B} &Random & 57.0 &15.8 \\
&~+~\ourmethod &{\textbf{60.5}} {\scriptsize (+3.5)} &{\textbf{19.3}} {\scriptsize (+3.5)}\\
\bottomrule
\end{tabular}
\end{minipage}
\vspace{-3mm}
\end{table*}

% \paragraph{Context Window Size} To examine whether~\ourmethod can scale to longer context lengths, we train a 1B-parameter model with a 32K context length from scratch on the Fineweb-Pro corpus containing 100B tokens. As the target context window $L$ increases, we adjust $\alpha$ to $1/2$ so that the local attention window will reach 32k at 64B tokens. We evaluate the model on the popular RULER benchmark~\citep{hsieh2024ruler} for long-context evaluation. Table~\ref{tab:32k-ruler} shows that our model performs well on both short and long tasks. Compared with the baseline of constant 32k pretraining, \ourmethod actually trains on shorter contexts. This highlights a counterintuitive finding that naively training a model on long contexts data is not always optimal.  
\begin{wraptable}{R}{0.39\linewidth}
\small
\vspace{-5mm}
% \begin{tabular}{lrrrrrrr}\toprule
% &32k &24k &16k &8k &4k & \textbf{Avg. Standard} \\\midrule
% Random &17.4 &26.9 &36.6 &41.2 &50.4 &50.7 \\
% +~\ourmethod &{19.1} &{27.7} &{38.1} &{48.0} &{58.1} &54.3 \\
% IntraDoc &21.6 &30.4 &41.1 &46.5 &57.7 &54.0 \\
% +~\ourmethod &{22.2} &{31.9} &{47.6} &{53.0} &{61.2} &{54.9 }\\
% \bottomrule
% \end{tabular}
\caption{Performance (\%) of 1B models trained on 100B FineWeb-Pro tokens with a 32K context window.}
\begin{tabular}{lll}\toprule
\textbf{Method} & \textbf{Standard} & \textbf{Long} \\\midrule
Random &50.7 &9.7\\
+~\ourmethod &54.3 {\scriptsize (+3.6)} &13.5 {\scriptsize (+3.8)}\\
IntraDoc &54.0 &13.0 \\
+~\ourmethod &\textbf{54.9} {\scriptsize (+0.9)} &\textbf{14.4} {\scriptsize (+1.4)} \\
\bottomrule
\end{tabular}

% \qian{Remember to cue that, with SkyLadder, either for 8k and 32k, our standard avg. performance is similar, demontrating that our method should be the best one to keep standard avg.}
\label{tab:32k-ruler}
%\end{table}
\vspace{-2mm}
\end{wraptable}

\paragraph{Context Window Size.} To examine whether \ourmethod can effectively scale to longer context windows, we train 1B models with a 32K context window on 100B FineWeb-Pro tokens. We adjust $\alpha$ to $1/2$ to ensure that the final context window expands to 32K before the end of pretraining. As shown in Table~\ref{tab:32k-ruler}, our model demonstrates strong performance on both standard and long benchmarks. In addition, the performance difference of \ourmethod (0.9\%) between the 8K and 32K models is largely reduced compared with the baseline approach (1.8\%), which alleviates the performance degradation described in our earlier study. Notably, compared to the baseline Random approach, \ourmethod trains the model on progressively shorter contexts during earlier stages. This reveals a counterintuitive insight: naively training a model with a long context window is not always optimal, even if the model is evaluated on long contexts. In contrast, strategic scheduling of the context window during pretraining can yield better results.

%\begin{table}[tb]\centering

% \paragraph{Coding} We show that the strategy could be used to improve code pretraining as well. 

% \subsection{Long-Context Continual Pretraining}
% \textcolor{blue}{We might not keep this section as the result is not very promising. Instead, keep a long context training from scratch setup could be better. }
% We now demonstrate that the dynamic masking schedule can be applied to long-context continual pretraining. For long-context models, several recent studies have suggested that training on long sequences lead to performance degradation on common benchmarks such as HellaSwag~\citep{zellers2019hellaswag} or MMLU~\citep{hendrycksmeasuring, haonan'swork, Gao tianyu's work, Zhiyuan's work}: the model after long-context continual training performs worse on these benchmarks than the original base model. We therefore apply the dynamic masking strategy in long-context extension to mitigate the degradation issue.

% As shown in Table xxx, dynamic masking shows outstanding performance in long-context pretraining. It achieves similar performance on long-context benchmark, RULER~\citep{hsieh2024ruler}, while maintaining the same level of performance on short benchmarks. This highlights the effectiveness of our approach beyond pretrainig from scratch: as long as there is a change in the model's context window, mask scheduling is helpful. 

\subsection{Ablation Study}\label{subsec:ablation}

% We now investigate the effect of the different hyperparameters in the schedule of \ourmethod. To reduce computation cost, the default setup is to pretrain a 120M model on 100B CC tokens to reach $L=8k$.

We now examine the impact of hyperparameters in \ourmethod scheduling. To manage computational costs, we adopt a default setup of pretraining 120M models with 8K context on 100B CC tokens. 

\begin{figure}[tb]
    \centering
    \includegraphics[width=0.485\textwidth]{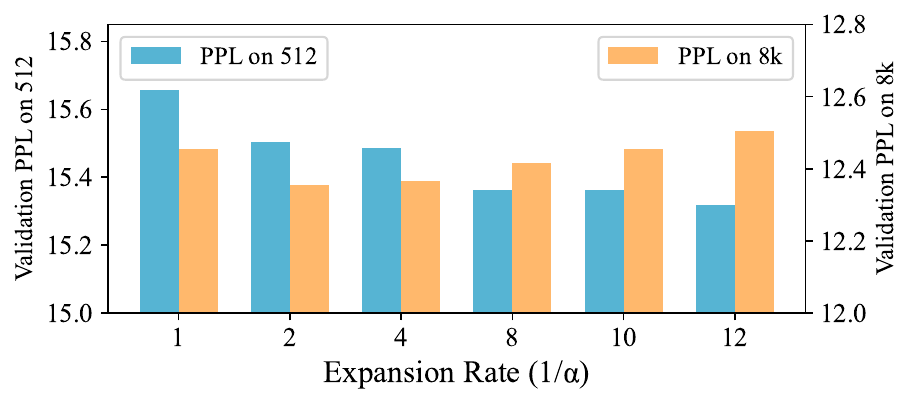} 
    \includegraphics[width=0.485\textwidth]{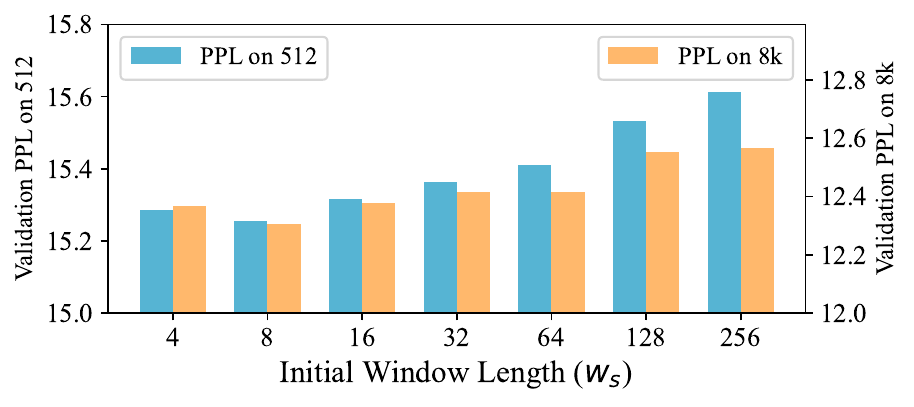} 
   \caption{Validation PPL on 512 and 8K contexts of models with different expansion rate $\alpha$ ({left}) and initial window length $w_s$ ({right}).}
    % \qian{@Tongyao: increase font size and use this tiny layout for the image...}}
    \label{fig:ppl-expansion-rate}
    \vspace{-1.5mm}
\end{figure}

\paragraph{Expansion Rate.} 
We investigate the impact of the expansion rate $\alpha$ in Figure~\ref{fig:ppl-expansion-rate} (left). We choose different $\alpha$ ranging from slowest ($1/12$) to fastest (1). Our findings reveal that, for short contexts, performance generally improves as the expansion rate slows down. 
However, selecting an excessively slow rate (e.g., $1/12$) can negatively affect long-context performance due to insufficient training on longer contexts.
\textbf{Therefore, we recommend setting $\alpha$ to $1/8$ for a good balance.}
% We investigate the impact of the expansion rate, $\alpha$, as illustrated in Figure~\ref{fig:expansion-rate-ablation}. We choose different $\alpha$ in ranging from slowest ($1/12$) to fastest (1). In Table~\ref{tab:expansion-rate-ablation}, we find that overall the performance on short contexts increases as the expansion rate slows down. However, choosing a value that is too slow (e.g. $1/12$) would harm long-context performance due to insufficient training on long contexts.  

\paragraph{Initial Context Window.}
As the final context window length $w_{e}$ is fixed to $L$, the sole remaining hyperparameter is $w_{s}$. Intuitively, setting $w_{s}$ to an excessively large value (e.g. close to $L$) leaves little room for scheduling, resulting in sub-optimal performance. In Figure~\ref{fig:ppl-expansion-rate} (right), we demonstrate that when $w_{s}$ is set to a relatively small value (e.g., 8), great performance can be achieved for both short and long contexts. This suggests that there is still potential for further improvement in our default setup. \textbf{Therefore, we recommend starting with a small context window, such as 8 tokens}.\looseness=-1

% As the final mask length, $m_{e}$ is set to $L$, the only remaining hyper-parameter is the $m_{s}$. Intuitively, setting $m_{s}$ to a very large value does not help to improve performance. In Figure~\ref{fig:initial-window-ablation}, we show that when $m_{s}$ is set to a very low value (e.g. 8), performance on both short and long contexts can be optimal. This shows that there is still room for our default setup. On the contrary, setting a value as large as 256 leads to a significant performance degradation. We therefore conclude that as long as $m_{s}$ is not too large, the final performance will not be significantly different.\looseness=-1

% \begin{figure}[t]
%     \centering
%     \includegraphics[width=0.48\textwidth]{figures/initial_window_ablation.pdf} 
%     \caption{Validation PPL on 512 and 8k contexts of models with different initial window length $w_s$.}
%     \label{fig:initial-window-ablation}
% \end{figure}

\paragraph{Scheduling Type.}
The default scheduling method in \ourmethod is linear scheduling. 
% Min: you said this earlier.  Omitting.
% Inspired by learning rate scheduling
We evaluate different context window scheduling types (more details in Table~\ref{tab:math-for-schedule-func} and Figure~\ref{fig:schedules_plot} in Appendix~\ref{app:sec:additional-ablation}): (1) \textbf{Stepwise Linear} rounds window size $w(t)$ to multiples of 1K, resulting in a step function;  (2) \textbf{Sinusoidal} increases quickly at the early stage then slows down; (3) \textbf{Exponential} starts slow but accelerates sharply; (4) \textbf{Continual pretraining} setup trains with 4K context windows for $\sim$97B tokens, then switches to 32K context for the final 3B tokens. Table~\ref{tab:32k-schedule-type} shows that linear and sinusoidal schedules outperform the exponential variant on long tasks, likely because the exponential schedule, with extended short-context pretraining at the beginning, fails to adequately train on long contexts. 
Last, the most commonly used continual pretraining setup performs poorly overall, suggesting abrupt context changes harm both short and long performance.
\textbf{These findings suggest that context window scheduling is superior to both constant long-context pretraining and continual pretraining}.

\begin{table*}[bt]
\begin{minipage}[t]{0.483\linewidth}

\small
\caption{Comparison of 1B models trained with a 32K context window with different scheduling methods. Numbers are average accuracy (\%).}
\begin{tabular}{lll}\toprule
\textbf{Method} &\textbf{Long} & \textbf{Standard} \\\midrule
Constant Long (32K) & 9.7 &50.7 \\
\midrule
Linear (32$\rightarrow$32K, \textit{default}) &13.5&54.3 \\
Stepwise Linear (32$\rightarrow$32K) &13.3 &\textbf{55.3} \\
Sinusoidal (32$\rightarrow$32K) &\textbf{14.2}&54.2 \\
Exponential (32$\rightarrow$32K) &11.5 &54.7 \\
Cont. Pretrain (4K$\rightarrow$32K) &10.0 &52.9 \\
\bottomrule
\end{tabular}
\label{tab:32k-schedule-type}
\end{minipage}
\quad
\begin{minipage}[t]{0.486\linewidth}
\caption{Comparison of relative training time and compute efficiency for 1B Models with different context window sizes $L$. FLOPs calculation follows~\citet{zhang2024tinyllamaopensourcesmalllanguage}. A larger context window leads to more efficiency gains. }
\small
\begin{tabular}{lll}\toprule
 \textbf{Method}  & \textbf{Time} (\%) & \textbf{FLOPs} ({$10^{20}$})\\\midrule
 Random (8K) & 100.0\% & 11.6  \\
 ~+~\ourmethod & 86.9\%~\scriptsize\textcolor{darkgreen}{(-13.1\%)} & 9.9~\scriptsize\textcolor{darkgreen}{(-14.7\%)}\\
\midrule
 Random (32K)   & 100.0\% & 25.5 \\
 ~+~\ourmethod & 77.8\%~\scriptsize\textcolor{darkgreen}{(-22.2\%)} & 18.8~\scriptsize\textcolor{darkgreen}{(-26.3\%)}\\
\bottomrule
\end{tabular}
\label{tab:training-efficiency-simplified}

\end{minipage}

% \begin{tabular}{lccccccc}\toprule
% Method &32k &24k &16k &8k &4k & \textbf{Standard Avg.} \\\midrule
% Random &17.4 &26.9 &36.6 &41.2 &50.4 &50.7 \\
% \midrule
% Linear (Default) &19.1 &\textbf{27.7} &38.1 &48.0 &58.1 &54.3 \\
% Linear Rounding &\textbf{19.4} &25.3 &41.2 &\textbf{49.1} &\textbf{62.9} &\textbf{55.3} \\
% Sinusoidal &18.6 &26.6 &\textbf{43.1} &48.5 &60.9 &54.2 \\
% Exponential &11.8 &17.6 &25.0 &42.3 &56.4 &54.7 \\
% Cont. Pretrain &7.7 &9.4 &16.0 &33.5 &50.3 &52.9 \\
% \bottomrule
% \end{tabular}
\vspace{-4mm}
\end{table*}

Overall, we conclude that the schedule should start from a small $w_s$ and the expansion should be gradual. We leave it to future work to study more advanced schedules and discover optimal configurations. For instance, it is possible that the schedule needs to be adjusted for various model sizes. More ablations for combination with BM25, hybrid attention, cyclic schedules and scheduling under a compute budget can be found in Appendix~\ref{app:sec:additional-ablation}.

% \begin{table}[tb]\centering
% \small
% \caption{Comparison of training time efficiency for 1B Models with different context window sizes.}
% \begin{tabular}{cll}\toprule
% \textbf{Context} & \textbf{Model} & \textbf{Relative Train Time} (\%) \\\midrule
% 8k & Random   & 100.0\%  \\
% 8k & ~+~\ourmethod & 86.9\%~~\textcolor{Green}{(-13.1\%)}\\
% \midrule
% 32k & Random   & 100.0\% \\
% 32k & ~+~\ourmethod & 77.8\%~~\textcolor{Green}{(-22.2\%)}\\
% \bottomrule
% \end{tabular}
% \label{tab:training-efficiency-simplified}
% \end{table}

\subsection{Analysis and Discussion}

% We observe a significant reduction in training time. In Table~\ref{tab:training-efficiency-simplified}, we show the total number of hours needed to train a baseline model with a constant context length, as well as a model with a schedule. We can see that we are able to save on average around 9-15\% total time because of the reduced context length overall on 8k sequences. On 32k context windows, the saving increases to 22\%. Therefore, if we consider the total compute or training time as a budget, \ourmethod allows for training more tokens and therefore the performance gain could be even larger. 
\paragraph{Training Efficiency.} We observe a significant boost in training efficiency when employing \ourmethod in Table~\ref{tab:training-efficiency-simplified}. On 8K models, \ourmethod accelerates training time by 13\% due to the reduced context window in calculating attention. With a 32K context window, the efficiency gain becomes even more pronounced: our method saves 22\% of training time while achieving better performance. The FLOPs saving is larger than the actual time because of reduced attention calculation.

%\subsection{Analysis}
% \qian{I would recommend move analysis to Section 4 and shorten its length, just leaving the attention pattern analaysis.}

\begin{wrapfigure}{r}{0.40\textwidth}
\vspace{-6mm}
    \centering
    \includegraphics[width=0.40\textwidth]{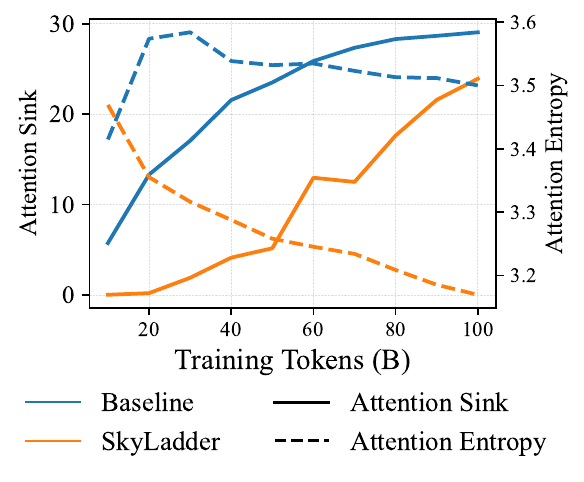} 
    \caption{Dynamics of attention sink and entropy during pretraining 1B models (8K context). \ourmethod delays the emergence of attention sink while lowering the overall entropy, indicating a more effective attention pattern. }
    \vspace{-2mm}
    \label{fig:sink-and-entropy}
\end{wrapfigure}

\paragraph{Attention Pattern.} We next investigate why \ourmethod, despite being trained on short contexts overall, consistently outperforms the baseline. As language models rely on attention mechanisms to encode context information, we study how attention patterns change. Specifically, during pretraining, we monitor the dynamics of (i) attention entropy (solid lines in Figure~\ref{fig:sink-and-entropy}), where a lower entropy is associated with better downstream performance~\citep{zhang2024attentionentropykeyfactor};
(ii) attention sink~\citep{xiao2024efficient}, where the initial token in the context receives disproportionately high attention. We utilize the metric in \citet{gu2024attentionsinkemergeslanguage} to quantitatively measure the amplitude of attention sink. As shown in Figure~\ref{fig:sink-and-entropy} (dashed lines), compared with the baseline Random, \ourmethod demonstrates reduced attention entropy, suggesting a more concentrated attention pattern. However, a slower emergence and lower amplitude of attention sink are simultaneously observed. 
This suggests that \ourmethod's attention is concentrated on the key information in the context instead of the initial token, which accounts for the performance gain. 

\paragraph{Training Stability.} 
\begin{wrapfigure}{r}{0.40\textwidth}
\vspace{-8mm}
    \centering
    \includegraphics[width=0.40\textwidth]{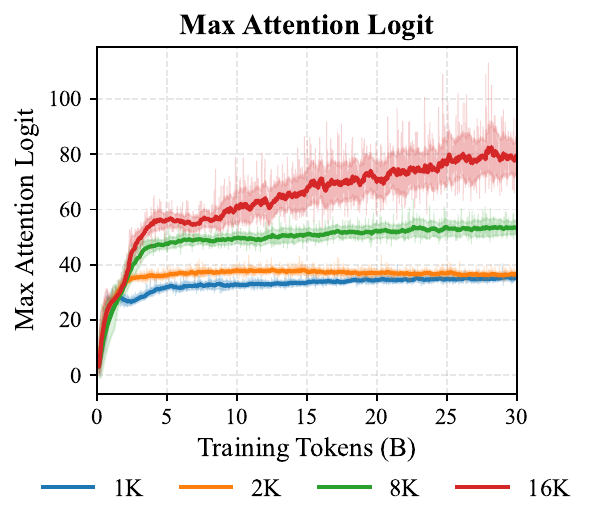} 
    \vspace{-6mm}
    \caption{Max attention logits during training of models of different context lengths (in different colors). }
    \label{fig:max-attention-logit}
    \vspace{-2mm}
\end{wrapfigure}

To further understand the reasons behind SkyLadder's better performance, we analyze the impact of pretraining context length on training dynamics. We pretrain 120M-parameter models with different context lengths. We first monitor the maximum attention logits ($S_{\text{max}} = \max_{i,j} \; q_i \cdot k_j$ for all $i,j$) throughout pretraining, following the methodology of K2~\citep{kimiteam2025kimik2openagentic}. A large attention logit indicates that an attention head is malfunctioning and may cause numerical instability. In Figure~\ref{fig:max-attention-logit}, we observe that pretraining with a long context of 16K tokens leads to exploding max attention logits, while a shorter window leads to lower attention logits. 

Next, we study the loss and gradient behavior by computing four stability metrics over the first $N = 30\mathrm{K}$ steps of pretraining, 
where $L_t$ denotes the training loss and $G_t$ is the gradient norm before clipping:

\begin{itemize}
    \item \textit{Loss Volatility:} measures local fluctuations of loss over a sliding window ($w = 10$), computed as $\frac{1}{N}\sum_{t=1}^{N}\mathrm{Std}(L_{t-w+1}, \ldots, L_t)$. Lower values indicate more stable training.
    \item \textit{Loss Smoothness:} the average loss change between consecutive steps, $\frac{1}{N-1}\sum_{t=2}^{N}|L_t - L_{t-1}|$. Smaller values mean smoother convergence.
    \item \textit{Mean Loss Ratio}~\citep{li2022stabilityefficiencydilemmainvestigatingsequence}: measures temporary increases in loss relative to the best loss so far, $\frac{1}{N-1}\sum_{t=2}^{N}\frac{L_t}{\min(L_{1}, \ldots, L_{t-1})}$, where smaller values indicate fewer loss spikes.
    \item \textit{Average Gradient Norm:} $\frac{1}{N}\sum_{t=1}^{N}\min(G_t, 1)$, where larger values indicate more aggressive gradient updates.
\end{itemize}

\begin{table*}[t]
\centering
\small
\caption{Training stability metrics during pretraining of 120M models with different context lengths. 
All metrics are averaged over the first 30 billion tokens. 
$\downarrow$ indicates that smaller values are better.}
\label{tab:stability-metrics}
\begin{tabular}{lcccc}
\toprule
Context & Volatility $\downarrow$ ($w{=}10$) & Smoothness $\downarrow$ & Mean Loss Ratio $\downarrow$ & Avg Grad Norm $\downarrow$ \\
\midrule
1K  & 0.023 & 0.019 & 1.014 & 0.335 \\
2K  & 0.026 & 0.023 & 1.017 & 0.338 \\
4K  & 0.030 & 0.029 & 1.020 & 0.340 \\
8K  & 0.036 & 0.036 & 1.025 & 0.347 \\
16K & 0.041 & 0.042 & 1.036 & 0.416 \\
\bottomrule
\end{tabular}
\end{table*}

In Table~\ref{tab:stability-metrics}, longer-context models show higher volatility, less smooth loss curves, more frequent upward spikes, and larger gradient norms, all indicating less stable optimization. In contrast, short-context models converge more smoothly with smaller fluctuations and more controlled gradient updates. Together, these results reveal that short-context pretraining is inherently more stable, both in attention behavior and optimization dynamics. The reduced numerical instability and smoother convergence likely enable more consistent gradient signals and better overall convergence, explaining their superior downstream performance.

\begin{wraptable}{R}{0.455\linewidth}
\vspace{-4.5mm}
\caption{Comparison between \ourmethod and Dataset Decomposition (DD) on 1B models trained with 100B FineWeb-Pro tokens. Numbers are in average performance in \%.}
\small
\begin{tabular}{lll}\toprule
\textbf{Model }&\textbf{Standard} &\textbf{Long} \\\midrule
IntraDoc & 54.3 & 12.7 \\
+ \ourmethod &\textbf{54.8} {\scriptsize (+0.5)} &\textbf{13.9} {\scriptsize (+1.2)} \\ 
+ DD (1 cycle) &53.9 {\scriptsize (-0.4)} &12.3 {\scriptsize (-0.4)} \\
+ DD (8 cycles) &54.5 {\scriptsize (+0.2)}&13.5 {\scriptsize (+0.8)}\\
\bottomrule
\end{tabular}
\label{tab:compare-related-work}
%\end{table}
\end{wraptable}
% \tongyao{I think this overlaps with related work. WE can consider merging them or shortening this part. }

% Our method is complementary to existing work on designing better pretraining lengths. For instance, \citet{li2022stabilityefficiencydilemmainvestigatingsequence} find that employing shorter sequence lengths as a warmup stage helps to stabilize training and produces lower loss. We further extend their conclusion by finding that the benefits of short context training are not limited to the warmup stage, but should be scheduled. On the other hand, 
\paragraph{Comparison with Related Work.} We compare our method with another approach for improving pretraining in Table~\ref{tab:compare-related-work}. As discussed in Section~\ref{sec:related_work}, \citet{pouransari2024dataset} proposed Dataset Decomposition~(DD) by segmenting a document into sequences of varying lengths and using a curriculum during pretraining. However, this approach inevitably introduces domain bias, as the document lengths in different domains are different~\citep{fu2024dataengineeringscalinglanguage}. This explains why DD with only one short-to-long cycle fails to outperform the IntraDoc baseline. To mitigate this, the authors suggested iterating through multiple cycles of long and short data, which does improve performance substantially. 
In contrast, our method achieves better performance by avoiding such biases by not altering the data order based on length. In Appendix~\ref{app:sec:additional-ablation}, we experimented with various cyclic schedules but did not observe any improvements. In fact, we noticed loss spikes between cycles (Figure~\ref{fig:schedule-cycle-ablation}), indicating potential issues with domain shifts. This further supports that our method is safer since it does not disrupt the natural ordering and distribution of the data. More discussion with other related works~\citep{li2022stabilityefficiencydilemmainvestigatingsequence, jin2023growlengthacceleratingllmspretraining} is in Section~\ref{app:sec:additional-compared-with-related-work}, where we demonstrate that our work provides novel insights that scheduling the context window over the entire training time improves both efficiency and performance. 
\section{Conclusion}
We conduct a comprehensive controlled study of the impact of context window on pretraining, revealing that a shorter context window is more beneficial to the model's performance on standard benchmarks. This debunks the trend of pretraining with longer context windows. We therefore propose \ourmethod to schedule the context window from short to long during pretraining, which gives substantial improvement in downstream performance and computational efficiency. We conclude that context window scheduling is an important dimension for pretraining, and deserves more consideration. In the future, we plan to explore more dynamic and performant scheduling strategies that adapt according to model size or pretraining data distribution.

%It is also interesting to study the interplay between other schedules (e.g. learning rate) and context window scheduling. 

% \qian{@Tongyao: remember to modify the conclusion accordingly, especially the method name. And please check the submission policy of ICML and whether we should have a limitation section?
% And please check the completeness of Appendix - if they are not well ready, remove them now.}

% \section*{Impact Statements}
% This paper presents work whose goal is to advance the field of Machine Learning. Our work enables the higher efficiency of general pretraining of Large Language Models. There are many potential societal consequences of our work, none of which we feel must be specifically highlighted here. 

% Bibliography entries for the entire Anthology, followed by custom entries
%\bibliography{anthology,custom}
% Custom bibliography entries only
%\bibliographystyle{iclr2025}

% \section*{Acknowledgement}
% We acknowledge the helpful discussion with Dou Longxu, and the feedback from members of NUS WING lab. Tongyao Zhu is supported by the EDB-IPP scholarship under Sea AI Lab. 
\newpage
\bibliography{custom}
\bibliographystyle{plainnat}

\newpage
\section*{NeurIPS Paper Checklist}

\begin{enumerate}

\item {\bf Claims}
    \item[] Question: Do the main claims made in the abstract and introduction accurately reflect the paper's contributions and scope?
    \item[] Answer: \answerYes{} % Replace by \answerYes{}, \answerNo{}, or \answerNA{}.
    \item[] Justification: Our abstract summarizes the findings of our preliminary study and the essence of the \ourmethod method: we systematically study the impact of context length on pretraining and empirically verify the effectiveness of our approach.  

\item {\bf Limitations}
    \item[] Question: Does the paper discuss the limitations of the work performed by the authors?
    \item[] Answer: \answerYes{} % Replace by \answerYes{}, \answerNo{}, or \answerNA{}.
    \item[] Justification: We discuss limitations of our study in Section~\ref{app:sec:limitation}.

\item {\bf Theory assumptions and proofs}
    \item[] Question: For each theoretical result, does the paper provide the full set of assumptions and a complete (and correct) proof?
    \item[] Answer: \answerNA{} % Replace by \answerYes{}, \answerNo{}, or \answerNA{}.
    \item[] Justification: This paper is not a theory paper and we do not include theoretical results. 

    \item {\bf Experimental result reproducibility}
    \item[] Question: Does the paper fully disclose all the information needed to reproduce the main experimental results of the paper to the extent that it affects the main claims and/or conclusions of the paper (regardless of whether the code and data are provided or not)?
    \item[] Answer: \answerYes{} % Replace by \answerYes{}, \answerNo{}, or \answerNA{}.
    \item[] Justification: We provide detailed hyperparameter setting (Section~\ref{para:training_setup}, \ref{app:sec:training-config}), implementation pseudocode (\ref{app:sec:implementation}) and model configuration (\ref{app:sec:model-architecture}). We implement most of the experiments in TinyLlama, a popular public project for pretraining, which makes it highly reproducible.  

\item {\bf Open access to data and code}
    \item[] Question: Does the paper provide open access to the data and code, with sufficient instructions to faithfully reproduce the main experimental results, as described in supplemental material?
    \item[] Answer: \answerYes{} % Replace by \answerYes{}, \answerNo{}, or \answerNA{}.
    \item[] Justification: Our pretraining code is based on TinyLlama, an open-source pretraining framework. The data is based on open-source pretraining corpora like SlimPajama and Fineweb-Pro. We include the code in the supplementary materials, and will open source the code upon acceptance. 

\item {\bf Experimental setting/details}
    \item[] Question: Does the paper specify all the training and test details (e.g., data splits, hyperparameters, how they were chosen, type of optimizer, etc.) necessary to understand the results?
    \item[] Answer: \answerYes{} % Replace by \answerYes{}, \answerNo{}, or \answerNA{}.
    \item[] Justification: We describe the training and testing setup in detail in Section~\ref{para:training_setup} for our preliminary study, and in Section~\ref{app:sec:sky-evaluation} for \ourmethod experiments. 

\item {\bf Experiment statistical significance}
    \item[] Question: Does the paper report error bars suitably and correctly defined or other appropriate information about the statistical significance of the experiments?
    \item[] Answer: \answerYes{} % Replace by \answerYes{}, \answerNo{}, or \answerNA{}.
    \item[] Justification: We conduct McNemar test for our main results in Table~\ref{tab:standard-tasks-1b-cc-model-performance}. The details for testing are in~\ref{app:sec:sky-evaluation}. Due to the excessive cost of pretraining, we do not perform repeated runs of the same setup. However, our claims are supported by multiple variations of the pretraining and comprehensive ablation studies in the main text and appendix. 

\item {\bf Experiments compute resources}
    \item[] Question: For each experiment, does the paper provide sufficient information on the computer resources (type of compute workers, memory, time of execution) needed to reproduce the experiments?
    \item[] Answer: \answerYes{} % Replace by \answerYes{}, \answerNo{}, or \answerNA{}.
    \item[] Justification: We share the compute information in Section~\ref{app:sec:compute-info}.
    
\item {\bf Code of ethics}
    \item[] Question: Does the research conducted in the paper conform, in every respect, with the NeurIPS Code of Ethics \url{https://neurips.cc/public/EthicsGuidelines}?
    \item[] Answer: \answerYes{} % Replace by \answerYes{}, \answerNo{}, or \answerNA{}.
    \item[] Justification: We fully anonymize the paper and the code. All experiments fully conform with the Code of Ethics. 

\item {\bf Broader impacts}
    \item[] Question: Does the paper discuss both potential positive societal impacts and negative societal impacts of the work performed?
    \item[] Answer: \answerYes{} % Replace by \answerYes{}, \answerNo{}, or \answerNA{}.
    \item[] Justification: We discuss the broader impact of our work in Section~\ref{app:sec:broader-impact}.
    
\item {\bf Safeguards}
    \item[] Question: Does the paper describe safeguards that have been put in place for responsible release of data or models that have a high risk for misuse (e.g., pretrained language models, image generators, or scraped datasets)?
    \item[] Answer: \answerNA{} % Replace by \answerYes{}, \answerNo{}, or \answerNA{}.
    \item[] Justification: This work is not meant to release a pretrained language model or publish a dataset. 

\item {\bf Licenses for existing assets}
    \item[] Question: Are the creators or original owners of assets (e.g., code, data, models), used in the paper, properly credited and are the license and terms of use explicitly mentioned and properly respected?
    \item[] Answer: \answerYes{} % Replace by \answerYes{}, \answerNo{}, or \answerNA{}.
    \item[] Justification: We explicitly mention and cite the source of the datasets (SlimPajama, FineWeb-Pro) and the implementation codebase in our paper (TinyLlama). License information is described in Section~\ref{app:sec:license}. All of them are open-source projects available for public use. 

\item {\bf New assets}
    \item[] Question: Are new assets introduced in the paper well documented and is the documentation provided alongside the assets?
    \item[] Answer: \answerYes{} % Replace by \answerYes{}, \answerNo{}, or \answerNA{}.
    \item[] Justification: We provide instructions to reproduce our experiments with the code in the supplementary materials. 

\item {\bf Crowdsourcing and research with human subjects}
    \item[] Question: For crowdsourcing experiments and research with human subjects, does the paper include the full text of instructions given to participants and screenshots, if applicable, as well as details about compensation (if any)? 
    \item[] Answer: \answerNA{} % Replace by \answerYes{}, \answerNo{}, or \answerNA{}.
    \item[] Justification: This project does not involve human subjects or crowdsourcing. 

\item {\bf Institutional review board (IRB) approvals or equivalent for research with human subjects}
    \item[] Question: Does the paper describe potential risks incurred by study participants, whether such risks were disclosed to the subjects, and whether Institutional Review Board (IRB) approvals (or an equivalent approval/review based on the requirements of your country or institution) were obtained?
    \item[] Answer: \answerNA{} % Replace by \answerYes{}, \answerNo{}, or \answerNA{}.
    \item[] Justification: This project does not involve human subjects or crowdsourcing. 

\item {\bf Declaration of LLM usage}
    \item[] Question: Does the paper describe the usage of LLMs if it is an important, original, or non-standard component of the core methods in this research? Note that if the LLM is used only for writing, editing, or formatting purposes and does not impact the core methodology, scientific rigorousness, or originality of the research, declaration is not required.
    %this research? 
    \item[] Answer: \answerNA{} % Replace by \answerYes{}, \answerNo{}, or \answerNA{}.
    \item[] Justification: We did not use LLMs for developing our method. 

\end{enumerate}

\newpage
\clearpage
\appendix

\section{Appendix}

\subsection{Limitations}\label{app:sec:limitation}
While we perform extensive experiments to study the impact of context window on pretraining and demonstrate the effectiveness of \ourmethod, we acknowledge that there are still limitations to be addressed. First, we conduct experiments up to a 3B-model scale and 32K context length, while the latest large language models are typically much larger and capable of processing longer contexts. However, pretraining a large model with a long context window requires prohibitive computational resources beyond our budget. Within our computational capabilities, we have tried to demonstrate the generalizability of \ourmethod across corpora, context window size, model size, and downstream tasks. Thus, we leave it as future work to apply \ourmethod to larger models. Second, we do not include a theoretical analysis to explain the effectiveness of \ourmethod as we mainly focus on empirical insights. We suggest that future work may investigate the relationship between the context window and the training compute to obtain the optimal context window schedule. 

\subsection{Broader Impacts}\label{app:sec:broader-impact}

The work aims to investigate the impact of choices on context windows in language model pretraining and proposes a way to speed up pretraining by scheduling context windows. On the positive side, this improves the efficiency of language model pretraining, making it more accessible and reducing the carbon footprint. Moreover, it enhances the performance of pretrained language models, which may result in better downstream performance in applications. There might be potential misuse of pretrained language models, which is beyond the scope of this work. 

\subsection{Model Architecture} \label{app:sec:model-architecture}
In Table~\ref{app:tab:model_architecture}, we list the architecture choices of the models trained, including the 120M, 360M, and 1B models based on the TinyLlama architecture~\citep{zhang2024tinyllamaopensourcesmalllanguage}. The 3B model is based on Llama3.2  architecture~\citep{dubey2024llama}. 

\begin{table*}[bt]
\centering
\small
\caption{Model configurations for pretrained language models. }\label{app:tab:model_architecture}
\begin{tabular}{@{}lcccc@{}}
\toprule
\textbf{Model} & \texttt{Tinyllama 1B} & \texttt{Tinyllama 120M} & \texttt{Tinyllama 360M} & \texttt{Llama3.2 3B} \\
\midrule
\textbf{Vocab Size} & 32000 & 32000 & 32000 & 32000 \\
\textbf{Layers} & 22 & 12 & 18 & 28 \\
\textbf{Heads} & 32 & 12 & 16 & 24 \\
\textbf{Embedding Dim} & 2048 & 768 & 1024 & 3072 \\
\textbf{Intermediate Size} & 5632 & 2048 & 4096 & 8192 \\
\textbf{Normalization} & RMSNorm & RMSNorm & RMSNorm & RMSNorm \\
\textbf{Normalization $\epsilon$} & $1\times10^{-5}$ & $1\times10^{-5}$ & $1\times10^{-5}$ & $1\times10^{-5}$ \\
\textbf{Query Groups} & 4 & 1 & 16 & 8 \\
\textbf{Bias} & No & No & No & No \\
\textbf{RoPE $\theta$} & \makecell[c]{10000 if $L=8K$ \\ 1000000 if $L=32K$} & 10000 & 10000 & 100000 \\
\bottomrule
\end{tabular}

\end{table*}

\subsection{Training Configurations}\label{app:sec:training-config}
We include details of the training configurations in Table~\ref{tab:training_config}. All models, irrespective of size or context window length, are trained on this same set of hyperparameters. For most of the hyperparameter values, we follow the TinyLlama~\citep{zhang2024tinyllamaopensourcesmalllanguage} project, therefore, our results are highly reproducible. 
\begin{table*}[bt]
\centering
\caption{Hyperparameters setup for pretraining the language models. All pretrained models follow the same structure. }
\begin{tabular}{cc}
\toprule
\textbf{Parameter}                 & \textbf{Value}                      \\ \midrule
Optimizer                          & AdamW                               \\
AdamW-$\beta_1$                    & 0.9                                 \\
AdamW-$\beta_2$                    & 0.95                                \\
Learning Rate Schedule             & Cosine                              \\
Peak Learning Rate                 & 4e-4                                \\
Minimum Learning Rate              & 4e-5                                \\
Warmup Steps                       & 2000                                \\
Gradient Norm Clipping             & 1                                   \\
Total Steps                        & 100,000                             \\
Global Batch Size                  & 1,048,576 ($2^{20}$) tokens         \\
Weight Decay                       & 0.1                                 \\ \bottomrule
\end{tabular}
\label{tab:training_config}
\end{table*}

% \subsection{Hardware and Compute}
% We pretrain all models from scratch with Nvidia A100 GPUs with either 40G or 80G memory. The training of the baseline 120M model (8k context) takes about 37 hours on a node with 8 GPUs. The 360M model takes around 80 hours, and the 1B model takes 200 hours. 

\subsection{Implementation}\label{app:sec:implementation}
We provide the pseudocode for implementing \ourmethod with Flash Attention 2~\citep{dao2023flashattention2}. The only change is to apply local causal masking with size $w$, and combine them with the original document boundaries under the IntraDoc scenario. It can easily be integrated into any model before calculating attention. The rest of the training pipeline remains unchanged.  

% \begin{tcolorbox}[colframe=lightblue, colback=white, coltitle=black, title=\ourmethod with Flash Attention 2]
% \begin{lstlisting}[language=Python, breaklines=true,basicstyle=\small]
% # Input: 
% # q,k,v: query, key, and value states (RoPE applied)
% # doc_boundaries: boundaries of documents (EOS token positions)
% # is_intradoc: whether the attention is intra-document or not
% # training_step: current training step
% # L: sequence length and final context window length

% w = min(L, get_current_mask_length(training_step))
% mask_boundaries = np.arange(w, L, w)  # [w, 2w, 3w, ...]

% if is_intradoc:
%     mask_boundaries = np.union1d(mask_boundaries, doc_boundaries)
    
% # max length between two boundaries
% max_seqlen = get_max_seqlen(mask_boundaries, L)  
% # cumulative sequence lengths, as required by flash attention
% cu_seqlens = get_cu_seqlens(mask_boundaries, L)  

% attn = flash_attn_varlen_func(q,k,v, cu_seqlens, max_seqlen, causal=True)
% \end{lstlisting}
% \end{tcolorbox}

\begin{tcolorbox}[
  enhanced,
  colframe=lightblue,
  colback=white,
  coltitle=black,
  fonttitle=\bfseries,
  title={\ourmethod{} with Flash Attention 2},
  boxrule=0.8pt,
  arc=2mm,
  top=1mm, bottom=1mm, left=1mm, right=1mm
]
\begin{lstlisting}[
  language=Python,
  basicstyle=\small\ttfamily,
  commentstyle=\color{gray}\ttfamily,
  keywordstyle=\color{blue}\bfseries,
  breaklines=true,
  columns=flexible,
  keepspaces=true,
  showstringspaces=false,
  frame=none
]
# q, k, v: RoPE-encoded query, key, value tensors
# doc_boundaries: EOS token positions per document
# is_intradoc: intra-document attention flag
# training_step: current global step
# L: maximum context window length

# get current window size
w = min(L, get_current_mask_length(training_step))

# breakpoints every w tokens (and at document boundaries if using IntraDoc masking)
mask_boundaries = np.arange(w, L, w)
if is_intradoc:
    mask_boundaries = np.union1d(mask_boundaries, doc_boundaries)

# compute max segment length & cumulative lengths for flash attention
max_seqlen = get_max_seqlen(mask_boundaries, L)
cu_seqlens = get_cu_seqlens(mask_boundaries, L)

attn = flash_attn_varlen_func(
    q, k, v,
    cu_seqlens,
    max_seqlen,
    causal=True
)
\end{lstlisting}
\end{tcolorbox}

\subsection{Definition of Per-token Context Window}\label{app:sec:definition-of-context}
In Figure~\ref{fig:packing-length-compare}(d), we show the context window distribution difference between IntraDoc and Random. To clarify, the context window size refers to the number of preceding tokens available in the context window when making the next token prediction. This is different from (a) and (b), where the context length $L$ is the model’s pretrained context window.

Formally, consider a token at index $i$ and an attention mask matrix $M$, where an entry $M_{i,j} = 0$ indicates that token $i$ can attend to token $j$, and $-\infty$ otherwise. The context window size $C_i$ for the $i$-th token is defined as $C_i = \sum_{j=1}^{i} \mathbf{1} \left\{M_{i,j} = 0\right\} $, where $\mathbf{1}\left\{\cdot\right\} $ is the indicator function that returns 1 when $M_{i,j} = 0$ and 0 otherwise. In essence, $C_i$ is the number of tokens available as context for the $i$-th token, and the distribution of $C_i$ over all pretraining tokens is in Figure~\ref{fig:packing-length-compare}(d).

For Random, the causal mask is triangular: the $i$-th token has a context window size equal to $i$ (i.e., $C_1 = 1$, $C_2 = 2$, etc.). Thus, the distribution of $C_i$ is uniform. In contrast, IntraDoc effectively shortens the context length by limiting the cross-document attention.

\subsection{Additional Results}

\subsubsection{Context Window Study}\label{app:context-length-study}

\begin{figure*}[htb]
    \centering
    \includegraphics[width=0.32\linewidth]{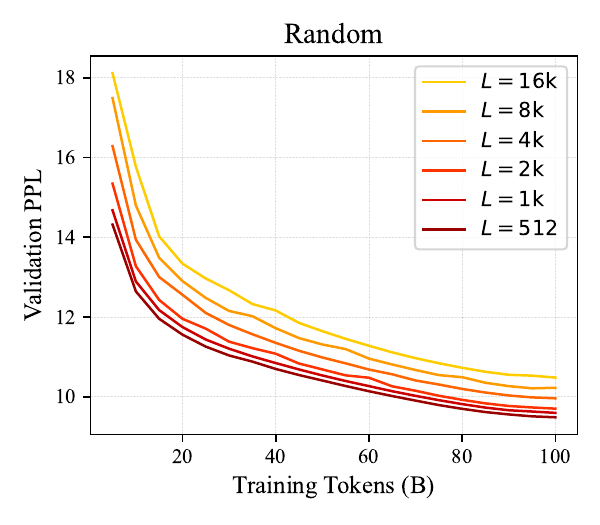}
    %\hspace{0.01\textwidth}
    \includegraphics[width=0.32\linewidth]{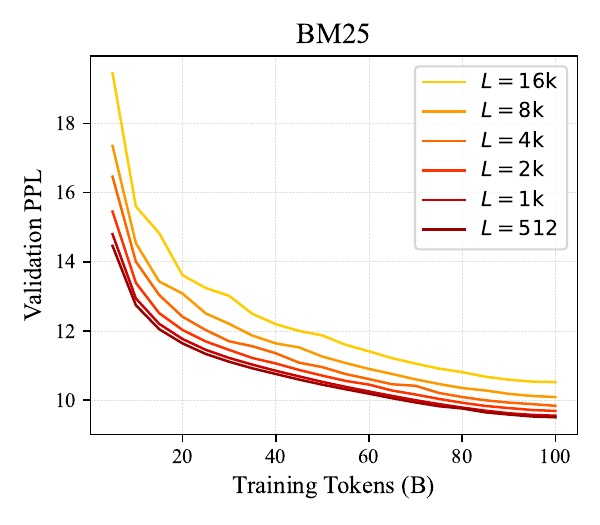}
    %\hspace{0.01\textwidth}
    \includegraphics[width=0.32\linewidth]{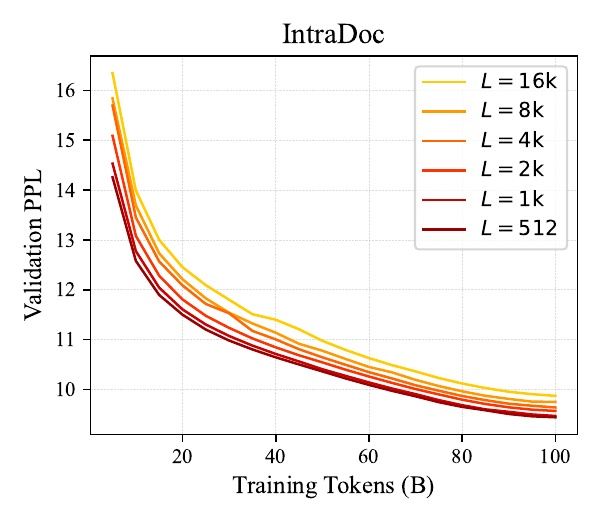}

    \caption{Validation perplexity (evaluated on a sliding window of 512) on models with different context lengths. }
    \label{fig:length-ppl-compare-1b}
\end{figure*}

\begin{figure}[htb]
%\begin{wrapfigure}{r}{0.5\linewidth}
\begin{minipage}[t]{0.5\linewidth}
    \centering
    \includegraphics[width=\textwidth]{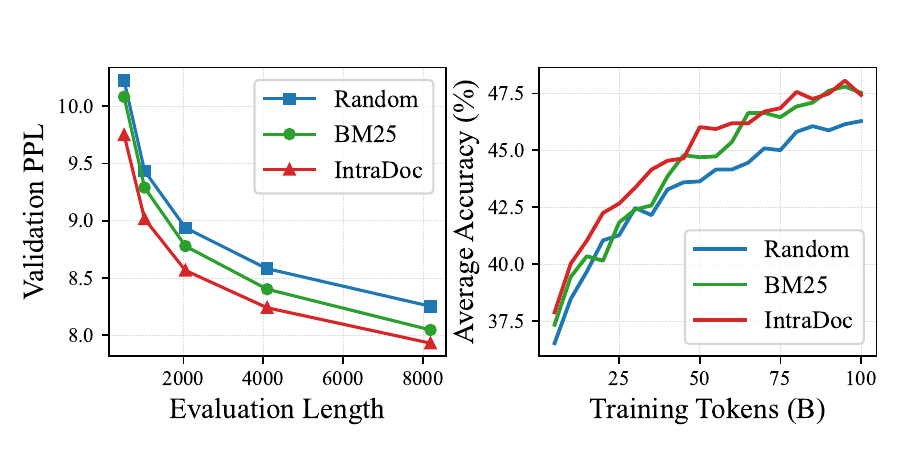}    
    \caption{Left: Evaluation perplexity of models with different packing or masking strategies. Right: Downstream performance over 9 tasks of different models. }
    \label{fig:packing-8k-perf}
%\end{figure}
\end{minipage}
    \quad\quad
    \begin{minipage}{0.43\linewidth}     
    \includegraphics[width = \textwidth]{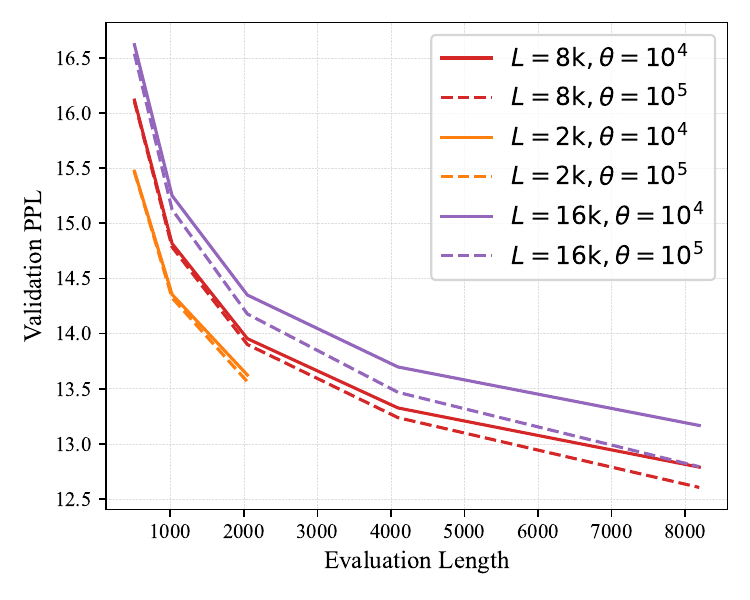} 
    \caption{Validation perplexity vs training tokens with different context windows and base of RoPE, $\theta$. Evaluation is done on a sliding window of varying length (x-axis) on the validation documents. }
    \label{fig:rope-base-ablation}
    \end{minipage}
\end{figure}

In Figure~\ref{fig:length-ppl-compare-1b}. We plot the validation perplexity of models with different context windows under the Random, IntraDoc, and BM25 settings. We observe a consistent trend that a shorter-context model has lower evaluation perplexity on a shorter sequence under all settings.

In Figure~\ref{fig:packing-8k-perf}, we plot the evaluation perplexity and downstream performance of models with different packing or masking strategies. We conclude that overall, IntraDoc achieves the best performance, with a consistently lower PPL and a higher downstream accuracy. We think that this is partially due to the shorter context window that the IntraDoc model is trained on.

\subsubsection{Ablations for Context Window Study}
% To ensure that the difference between models with different contexts is significant, we study different settings with the positional encoding method, which is largely affected by changing the context length.

\paragraph{Base of RoPE.} It has been shown that the value of RoPE may have a significant impact on the model's long context performance, and a longer context requires a larger base~\citep{men2024baseropeboundscontext}. Therefore, we increase the RoPE base to 100,000, which is sufficiently large according to~\citet{men2024baseropeboundscontext}. In Figure~\ref{fig:rope-base-ablation}, we observe an improvement for long-context models on long-context evaluation. However, the large gap between a shorter and a longer model still remains, therefore rejecting the hypothesis that the RoPE base is the key contributing factor to the superior performance of short-context models. \label{app:rope-base-ablatoion}

% \paragraph{No Positional Embedding (NoPE)} 
% To further eliminate the impact of RoPE, we entirely remove the position encoding. The model should still be able to learn implicit position information~\citep{kazemnejad2023impactpositionalencodinglength}. Figure~\ref{fig:nope-compare-ablation} shows the performance of NoPE-based models across training steps. As expected, NoPE models underperforms RoPE due to loss of positional information. However, even under the NoPE setting, short-context models also outperforms long-context models, which highlights that our findings are not purely due to RoPE. 

% \begin{table*}[!htp]\centering
% \caption{Performance of 3B models. WG stands for WinoGrande and HS denotes HellaSwag. }\label{tab:3b-olmes}

% \begin{tabular}{lrrrrrrrrrrrr}\toprule
% Model &Avg. &ARC-E &ARC-C &BoolQ &CSQA &HS &OBQA &PIQA &SIQA &WG &MMLU \\\midrule
% intramask &59.7 &75.9 &45.5 &67.1 &63.1 &69.6 &48.2 &73.8 &51.4 &64.2 &38.4 \\
% intradm8 &60.7 &76.4 &47.8 &68.6 &65.2 &70.1 &49.2 &75.1 &52.6 &63.4 &38.9 \\
% random &58.0 &74.3 &43.3 &66.8 &63.2 &64.4 &46.4 &73.2 &49.4 &62.0 &36.7 \\
% dm8 &61.7 &78.1 &46.8 &72.7 &66.3 &70.3 &50.8 &75.1 &53.4 &64.3 &39.6 \\
% \bottomrule
% \end{tabular}
% \end{table*}

\subsubsection{\ourmethod Evaluation}\label{app:sec:sky-evaluation}
%\begin{table}[tb]\centering

%\end{table}
\paragraph{Statistical Test} We test the statistical significance of the performance difference between our models and baselines in Table~\ref{tab:standard-tasks-1b-cc-model-performance}. We use a McNemar test as the two models are evaluated on the same set of questions. The original OLMES suite samples 1000 examples from each benchmark's full evaluation suite. In contrast, when conducting the McNemar test, we evaluate models on the full set to obtain more statistically meaningful results. We note that OpenBookQA only has 500 questions, making it harder to obtain statistical significance. 
\paragraph{Reading Comprehension}
For reading comprehension, we evaluate the following benchmarks: Hotpot QA (2-shot)~\citep{yang-etal-2018-hotpotqa}, SQuAD (4-shot)~\citep{rajpurkar-etal-2016-squad}, NaturalQuestions (NQ) (2-shot)~\citep{kwiatkowski-etal-2019-natural}, TriviaQA (2-shot)~\citep{joshi-etal-2017-triviaqa}, and RACE-high (0-shot)~\citep{lai-etal-2017-race}. We follow the setup by~\citet{zhao2024analysing}, where NQ and TriviaQA use retrieved documents as contexts. For RACE, we use \texttt{lm-evaluation-harness}~\citep{eval-harness} to compare the PPL between options. 

\begin{table*}[tb]
\centering
\small
\caption{Performance of 3B models on long tasks of retrieval-augmented generation (evaluated by exact-match scores) and reading comprehension benchmarks (accuracy in \%). }
\label{tab:combined-results}
\begin{tabular}{l ccccc ccc}
\toprule

  & \multicolumn{5}{c}{\textbf{Retrieval Augmented Generation}}
  & \multicolumn{3}{c}{\textbf{Reading Comprehension}} \\
\cmidrule(lr){2-6} \cmidrule(lr){7-9}
 \textbf{Model}& Avg. & NQ & TriviaQA & HotpotQA & PopQA
 & Avg. & TOEFL & QuALITY \\
\midrule
Random    & 30.3 & 24.3 & 45.2 & 29.3 & 22.5 
            & 37.1 & 43.5 & 30.6 \\
+~SkyLadder & \textbf{35.5} & \textbf{27.8} &\textbf{52.7} & \textbf{32.3} & \textbf{29.3} 
            & \textbf{39.4} & \textbf{48.0} & \textbf{30.9} \\
\bottomrule
\end{tabular}
\label{app:tab:3b-model-reading-and-rag}
\end{table*}

\begin{table*}[tb]
\centering
\small
\caption{Many‐shot ICL performance (accuracy) on text classification benchmarks. Numbers in parentheses denote the number of labels for each task. }
\label{tab:manyshot-icl}
\begin{tabular}{lcccccc}
\toprule
\textbf{Model}
  & \textbf{Avg.}
  & \textbf{DBpedia (14)}
  & \textbf{AGNews (4)}
  & \textbf{Amazon (2)}
  & \textbf{Yelp (2)}
  & \textbf{SST2 (2)} \\
\midrule
Random
  & 73.9
  & 17.4
  & 68.6
  & \textbf{94.3}
  & 94.7
  & \textbf{94.5} \\
+~SkyLadder
  & \textbf{76.5}
  & \textbf{25.5}
  & \textbf{75.8}
  & 94.1
  & \textbf{95.0}
  & 92.2 \\
\bottomrule
\end{tabular}
\label{app:tab:3b-model-icl}
\end{table*}

\paragraph{Long-context Evaluation} We provide additional long-context evaluation on our largest 3B model with an 8K context. This is to mitigate the performance instability of using synthetic benchmarks on small models. We first follow~\citep{pouransari2024dataset} to evaluate model accuracy on reading comprehension benchmarks TOEFL~\citep{chung2018supervisedunsupervisedtransferlearning,tseng2016machinecomprehensionspokencontent} and QuALITY~\citep{pang-etal-2022-quality}. Next, we evaluate the model's performance on Retrieval Augmented Generation (RAG), where the model is provided with many relevant but potentially noisy contexts and needs to locate the correct information. As shown in Table~\ref{app:tab:3b-model-reading-and-rag}, \ourmethod consistently performs better than the baseline across all evaluated RAG and reading comprehension datasets, highlighting its ability to locate correct answers within a lengthy context. In addition, we test the in-context learning ability of the models on text classification benchmarks ~\citep{zhao2024analysing,shicontext}. Results in Table~\ref{app:tab:3b-model-icl} suggest that SkyLadder shows a significant gain for tasks with many labels, such as DBpedia, while achieving comparable high performance on binary tasks. 
\begin{wraptable}{r}{0.5\linewidth}
\small
\centering
\caption{1B model (trained on CC) performance (exact match \%) on closed-book QA tasks. }\label{tab:1b-cc-cbqa}
\begin{tabular}{lrrr}\toprule
&\multicolumn{3}{c}{\textbf{Closed-book QA}} \\\cmidrule(lr){2-4}
Model &NQ &TriviaQA &Average \\\midrule
Random &6.1 &11.9 &9.0 \\
+~\ourmethod &9.0 &17.5 &\textbf{13.2} \\
IntraDoc &7.8 &14.7 &11.3 \\
+~\ourmethod &8.2 &17.4 &\textbf{12.8} \\
\bottomrule
\end{tabular}
\vspace{-10mm}
\end{wraptable}

\vspace{-4mm}
\paragraph{Closed-book QA} We additionally evaluate the closed-book QA performance of our models without access to any document. We use the evaluation protocol ~\citet{zhao2024analysing} to measure the exact match. In Table~\ref{tab:1b-cc-cbqa}, we notice a significant improvement in our methods compared to the baselines for answering closed-book questions. This is consistent with the results that our models show improvements on standard benchmarks that contain commonsense knowledge.

\subsubsection{\ourmethod Ablations}\label{app:sec:additional-ablation}
% \begin{wraptable}{r}
% {0.5\linewidth}
% \small
\begin{table*}[t]
\begin{minipage}[t]{0.50\linewidth}
\centering
\small
\caption{Performance (\%) of 1B models with different schedule types. All models are trained on the same 100B CommonCrawl tokens with a final context length of 8K. BM25 packing, when combined with \ourmethod, significantly boosts performance on long tasks.}\label{tab:bm25-with-skyladder}
\begin{tabular}{lll}\toprule
&\textbf{Standard Avg.} &\textbf{Long Avg.} \\\midrule
Random &46.3 &15.3 \\
BM25 &47.5 \scriptsize{(+1.2)} &16.4 \scriptsize{(+1.1)} \\
+~\ourmethod &\textbf{49.8} \scriptsize{(+3.5)} &\textbf{17.0} \scriptsize{(+1.7)}\\
\bottomrule
\end{tabular}

\end{minipage}
\quad\quad
\begin{minipage}[t]{0.48\linewidth}
\small
\caption{Performance (\%) of 1B models with different schedule types. All models are trained on the same 100B FineWeb-Pro tokens with a final context length of 8K. Short-to-long scheduling is consistently better than long-to-short scheduling. }\label{tab:long-to-short}
\begin{tabular}{lll}\toprule
&\textbf{Standard Avg.} &\textbf{Long Avg.} \\\midrule
No Scheduling &52.5 &11.1 \\
Short-to-Long &\textbf{55.2} {\scriptsize (+2.7)} &\textbf{12.3 }{\scriptsize (+1.2)} \\
Long-to-Short &52.6 {\scriptsize (+0.1)} &10.7 {\scriptsize (-0.4)} \\
\bottomrule
\end{tabular}

\end{minipage}

\end{table*}
%\end{wraptable}

\paragraph{Combination with BM25 Packing} As \ourmethod only changes the context length via masking without altering the underlying data, it is orthogonal to any advanced data packing method such as~\citet{shicontext, ding2024fewertruncationsimprovelanguage}. In Table~\ref{tab:bm25-with-skyladder}, we combine the \ourmethod with the BM25 packing method. We show that the model achieves even better performance on both short and long context evaluation than BM25 without scheduling, which is also better than the Random baseline. This reveals that our method can be combined with more advanced packing techniques to further boost performance.

\paragraph{Combination with Hybrid Attention}
We note that a recent interesting trend in pretraining models with long context is to use a hybrid attention structure. For instance, Gemma3~\citep{gemmateam2025gemma3technicalreport} uses a mixture of global and local attention layers to balance efficiency and performance of the long-context model. We are curious about the generalizability of \ourmethod to such architecture, and follow Gemma3's strategy with a global-to-local ratio of 6:1. The results are presented in Table \ref{app:tab:gemma3-skyladder}. We observe that SkyLadder consistently outperforms the baseline across all evaluation lengths, verifying its applicability. Importantly, SkyLadder works along the time dimension and is combinable with different attention variants, as long as there is a context window to be scheduled. We also verified the effectiveness of SkyLadder on alternative model structures. In Table~\ref{app:tab:qwen0.5b-ppl}, we pretrain models following the Qwen2.5-0.5B structure, and obtain consistent gains as well. 
\begin{table*}[tb]
\centering
\small
\caption{Evaluation perplexity for Gemma3-like models with different evaluation context lengths $L_e$. All models are trained with 100B tokens on CommonCrawl.}
\label{app:tab:gemma3-skyladder}
\begin{tabular}{lccc}
\toprule
Model & $L_e=512$ & $L_e=4K$ & $L_e=8K$ \\
\midrule
Random -- 120M    & 15.9 & 13.4 & 13.0 \\
+ SkyLadder         & 15.5 & 12.9 & 12.4 \\
Random -- 360M    & 12.1 & 10.2 &  9.8 \\
+ SkyLadder         & \textbf{11.6} &  \textbf{9.7} &  \textbf{9.4} \\
\bottomrule
\end{tabular}
\end{table*}

\begin{table*}[tb]
\centering
\small
\caption{Evaluation perplexity for Qwen2.5-0.5B models under different evaluation context lengths $L_e$. All models are trained with 100B tokens on CommonCrawl.}
\label{app:tab:qwen0.5b-ppl}
\begin{tabular}{lccc}
\toprule
Model & $L_e=1$K & $L_e=4$K & $L_e=8$K \\
\midrule
Random         & 14.8 & 13.1 & 12.5 \\
+ SkyLadder    & 14.3 & 12.7 & 12.1 \\
\bottomrule
\end{tabular}
\end{table*}

\paragraph{Long-to-Short Schedule} A possibility that \ourmethod works better than baseline on standard benchmarks, which are typically short, might be that the training data mix has more short-context data after applying the mask.
To study the effect of pure data distribution, we conduct an ablation of reversing the original short-to-long schedule and name it as the long-to-short schedule. This schedule spends the same number of tokens (64B) in the changing phase, before the constant training phase in $L=8K$ for another 36B tokens. In Table~\ref{tab:long-to-short}, we show that the long-to-short schedule is not helpful to the model's performance in both short and long evaluation tasks. This highlights that the context window needs to be scheduled, rather than simply having a data mixture of long and short contexts.

\paragraph{Alternative Schedule Types} We explore various types of short-to-long scheduling following different functions as mentioned in Section~\ref{subsec:ablation}. Table \ref{tab:math-for-schedule-func} shows the details of the schedule as a function of $t$, and Figure ~\ref{fig:schedules_plot} shows an illustration of the different schedule types. In Table~\ref{tab:32k-schedule-type}, we show that a smoother increase following the sinusoidal schedule works the best for long-context evaluation, while also achieving strong performance on standard benchmarks.

\begin{table*}[tb]
    \begin{minipage}{0.48\textwidth}  % Adjust width as needed
    \small
        \caption{Functions for different context window schedule types. We set $w_s=32$ and $w_e=32768$ in our experiments. The $r$ for rounding is set to $1024$. }
    \label{tab:math-for-schedule-func}
    \begin{tabular}{ll}
        \toprule
        \textbf{Schedule} & \textbf{Function} \\
        \midrule
        Constant & $ w_e$ \\
        Linear & $w_s + (w_e - w_s) \frac{\alpha x}{w_e - w_s}$ \\
        Stepwise & $\max(w_s, r \times \left\lfloor \frac{L(x)}{r} \right\rfloor)$  \\
        Sinusoidal & $w_s + (w_e - w_s) \sin\left(\frac{\alpha\pi x}{2(w_e - w_s)} \right)$ \\
        Exponential & $w_s \times \left( \frac{w_e}{w_s} \right)^{\frac{\alpha x}{w_e - w_s}}$ \\
        \bottomrule
    \end{tabular}

    \end{minipage}
    \quad\quad
    \begin{minipage}{0.48\textwidth}  % Adjust width as needed
        \centering
        \includegraphics[width=\linewidth]{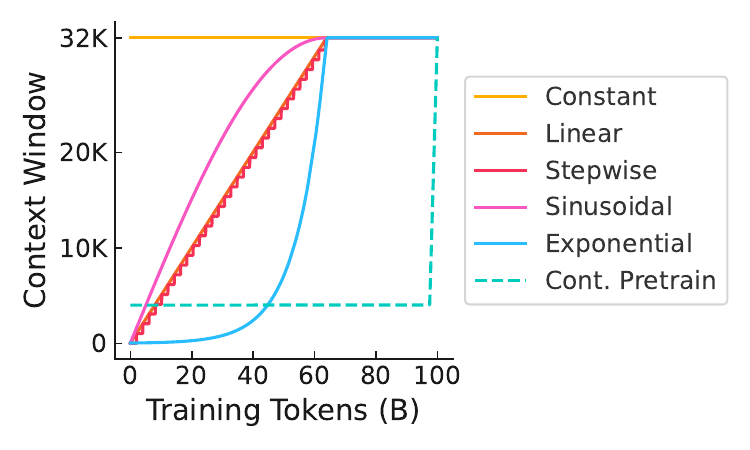}  % Adjust filename as needed
        \captionof{figure}{Illustration plot of various scheduling types. }
        \label{fig:schedules_plot}
    \end{minipage}
    
\end{table*}

% \begin{wraptable}{r}
% {0.5\linewidth}
% \small

% \end{wraptable}

% \paragraph{Continual Pretraining} 

\begin{figure}[hbt]
    \centering

    \begin{minipage}[tb]{0.48\linewidth}
    \centering
    \includegraphics[width = \textwidth]{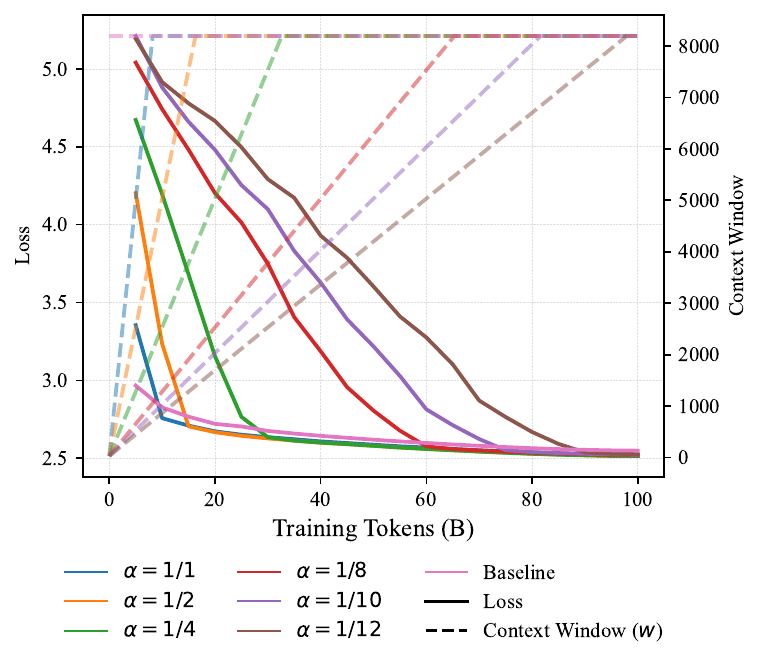} 
    \caption{An illustration of the effect of different $\alpha$. Dashed lines represent the current context window $w$ for each step, and solid lines are the loss evaluated at 8K length. }
    \label{fig:expansion-rate-ablation}
\end{minipage}\quad
\begin{minipage}{0.485\linewidth}
        \centering \vspace{2mm}
    \includegraphics[width = \textwidth]{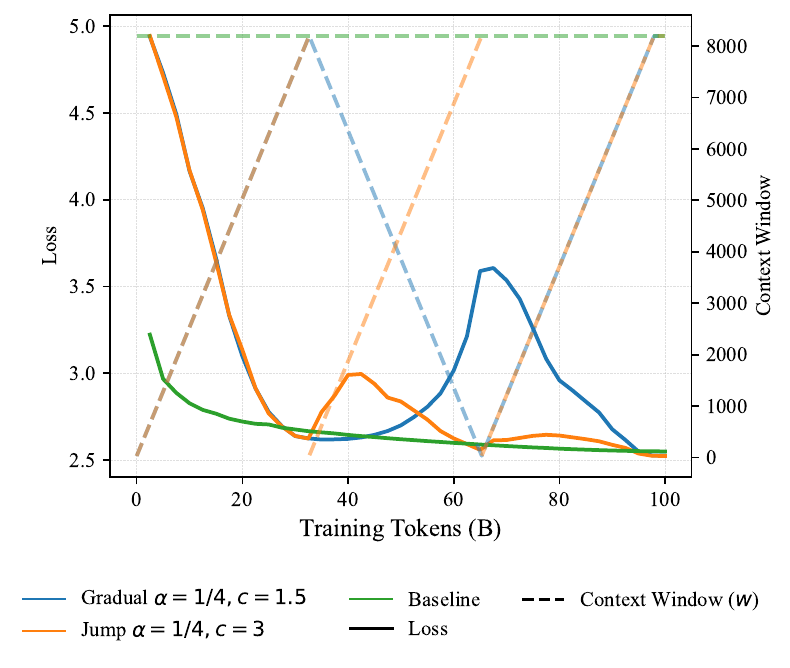}
    \caption{An illustration of the cyclic schedules with gradual increases or jumps. Dashed lines represent the context length for each step, and solid lines are the loss evaluated at an 8K length. $c$ represents the number of cycles.  }
    \label{fig:schedule-cycle-ablation}
\end{minipage}
\vspace{-3mm}
\end{figure}

\paragraph{Expansion Rate} We illustrate the effect of the rate of expansion $\alpha$ in Figure~\ref{fig:expansion-rate-ablation}. As the evaluation is done on 8K contexts, models with a lower rate (and shorter context window) will have a higher loss as the evaluation length is out-of-distribution. However, eventually, all models' loss converges to a low level after the schedule reaches 8K. The detailed numbers of validation loss after pretraining can be found in Table~\ref{tab:expansion-rate-ablation}. Following previous work~\citep{fu2024dataengineeringscalinglanguage, hoffmann2022training,kaplan2020scaling}, we consider a loss difference larger than 0.01 as significant. We conclude that setting a reasonable rate of $1/8$ balances both short and long-context loss, which is the default setup for our main experiments.

\begin{table*}[bt]\centering
\small

\caption{Validation loss with different expansion rates. A box is colored red if it is significantly worse (difference $>$ 0.01) than the best of the column. $L_e$ is the evaluation context length. All models are of size 120M and trained on 100B tokens. }\label{tab:expansion-rate-ablation}
\begin{tabular}{ccccc}
\toprule

\makecell{Rate\\($1/\alpha$)} &\makecell{Tokens to Reach 8K (B)} &$L_e = 512 $ &$L_e = 4K$ &$L_e = 8K$ \\\midrule
1 &8 &\cellcolor{red!22}2.751 &\cellcolor{red!12}2.563 &2.522 \\
2 &16 &\cellcolor{red!12}2.741 &2.551 &\textbf{2.514} \\
4 &32 &\cellcolor{red!11}2.740 &\textbf{2.551} &2.515 \\
8 &64 &2.732 &2.553 &2.519 \\
9 &72 &2.731 &2.553 &2.519 \\
10 &80 &2.732 &2.555 &2.522 \\
11 &88 &2.730 &2.554 &2.521 \\
12 &96 &\textbf{2.729} &2.557 &\cellcolor{red!12}2.526 \\
\multicolumn{2}{c}{Baseline (Constant)} &\cellcolor{red!51}2.780 &\cellcolor{red!39}2.590 &\cellcolor{red!35}2.549 \\
\bottomrule
\end{tabular}

\end{table*}

\paragraph{Cyclic Schedule} Inspired by the cyclic schedule learning rate~\citep{smith2017cyclical}, we also wonder if cycles are helpful in the schedule. In Figure~\ref{fig:schedule-cycle-ablation}, we show two cyclic schedules. In the ``Jump'' schedule, $w(t)$ will decrease to $w_s$ immediately after reaching $L$. On the other hand, the ``Gradual'' schedule means an ``M'' shape alternating between $w_e$ and $w_s$. Notably, in the discontinuous Jump schedule, we notice a significant increase in long-context perplexity when we train on only short contexts for an extended period. However, as long as $w$ increases back to $L$, the performance will return. 

In Table~\ref{tab:cycle-ablation}, we show that these schedules have no major impact on the final performance. This highlights that the method does not introduce additional bias in data selection: different from existing methods such as~\citet{pouransari2024dataset} that proposes to train on short data first, followed by long data, we do not assume such a curriculum on data. We argue that the context window size should be independent of the data lengths to avoid bias in training only on certain domains of data. 
% An illustration of the cyclic schedule can be found in Figure~\ref{fig:schedule-cycle-ablation}. In Table~\ref{tab:cycle-ablation}, we show the results of having multiple cycles in the schedule. We observe no significant improvement in enabling multiple cycles. 
% \begin{figure}[tb]

% \end{figure}

% \begin{table*}[tb]\centering
% \caption{Validation loss with cyclic schedules. $L_e$ represents the evaluation context length. }\label{tab:cycle-ablation}
% \begin{tabular}{lrrrrr}\toprule
% Type &\makecell{Number\\of\\Cycles} &\makecell{Tokens\\per\\Cycle (B)} &$L_e = 512 $ &$L_e = 4k$ &$L_e = 8k$ \\\midrule

% Baseline (Random) & & &2.780 &2.590 &2.549 \\
% +\ourmethod & & &2.732 &\textbf{2.553} &\textbf{2.519} \\\midrule
% Gradual &4.5 &16 &2.743 &2.565 &2.530 \\
% Jump &9 &8 &2.744 &2.566 &2.532 \\
% Gradual &2.5 &32 &2.732 &2.554 &2.521 \\
% Jump &5 &16 &2.733 &2.554 &2.521 \\
% Gradual &1.5 &64 &2.728 &2.556 &2.524 \\
% Jump &3 &32 &\textbf{2.727} &2.554 &2.522 \\
% \bottomrule
% \end{tabular}
% \end{table*}

\begin{table*}[bt]\centering

\small
\caption{Validation loss with cyclic schedules. $L_e$ represents the evaluation context length. All models are of size 120M and trained on 100B tokens. }\label{tab:cycle-ablation}
\begin{tabular}{lrrrr}\toprule
Type &\makecell{Number of Cycles} &\makecell{Tokens per Cycle (B)} &$L_e = 512 $ &$L_e = 8K$ \\\midrule

Random & & &2.780 &2.549 \\
+ \ourmethod & & &2.732 &\textbf{2.519} \\\midrule
Gradual &4.5 &16 &2.743 &2.530 \\
Jump &9 &8 &2.744 &2.532 \\
Gradual &2.5 &32 &2.732 &2.521 \\
Jump &5 &16 &2.733 &2.521 \\
Gradual &1.5 &64 &2.728 &2.524 \\
Jump &3 &32 &\textbf{2.727} &2.522 \\
\bottomrule
\end{tabular}

\end{table*}

\paragraph{Initial Window Length}
We show the effect of having different $w_s$, the initial window length when the training starts. In Table~\ref{tab:start-length-ablation-dm4}, we show that the optimal starting length is 8 tokens. The trend is the same across both $\alpha=1/4$ and $\alpha=1/8$. This suggests that the starting length should be sufficiently small, irrespective of the expansion rate. It also reveals that prior studies, such as~\citet{jin2023growlengthacceleratingllmspretraining} and~\citet{pouransari2024dataset} that start with an initial length of 256 could be suboptimal.

% \begin{table}[tb]\centering
% \caption{Final validation loss after training a 120M model on 100B tokens with different $w_{s}$ and $\alpha=1/8$. $L_e$ represents the context length of evaluation. A cell is colored red if its loss has a $>0.01$ difference than the column's best.}\label{tab:start-length-ablation-dm8}
% \begin{tabular}{lrrrr}\toprule
% $w_s$ &$L_e=512$ &$L_e=4k$ &$L_e=8k$ \\\midrule
% 4 &2.727 &2.549 &2.515 \\
% 8 &\textbf{2.725} &\textbf{2.545} &\textbf{2.510} \\
% 16 &{2.729} &2.550 &2.516 \\
% 32 &2.732 &2.553 &2.519 \\

% 64 &2.735 &2.553 &2.519 \\
% 128 &\cellcolor{red!18}2.743 &\cellcolor{red!19}2.564 &\cellcolor{red!20}2.530 \\
% 256 &\cellcolor{red!23}2.748 &\cellcolor{red!22}2.567 &\cellcolor{red!21}2.531 \\
% 8192 &\cellcolor{red!55}2.780 &\cellcolor{red!45}2.590 &\cellcolor{red!39}2.549 \\
% \bottomrule
% \end{tabular}
% \end{table}

\begin{table*}[tb]\centering
\small
\caption{Final validation loss after training 120M models on 100B tokens with different $w_{s}$ when $\alpha=1/4$ and $\alpha=1/8$. $L_e$ represents the context length of evaluation. A cell is colored red if its loss has a difference larger than 0.01 from the column's best. $w_s=8192$ equals no scheduling. }\label{tab:start-length-ablation-dm4}

\begin{tabular}{lrrrr}\toprule
$w_s$ &$L_e=512$ &$L_e=4K$ &$L_e=8K$ \\\midrule
\multicolumn{4}{c}{$\alpha=1/4$} \\
4 &2.731 &2.546 &2.510 \\
8 &\textbf{2.730} &\textbf{2.545} &\textbf{2.508} \\
16 &2.733 &{2.551} &{2.513} \\
32 &2.740 &{2.551} &2.515 \\
64 &\cellcolor{red!12}2.742 &\cellcolor{red!12}2.557 &\cellcolor{red!12}2.520 \\
128 &\cellcolor{red!18}2.748 &\cellcolor{red!19}2.564 &\cellcolor{red!20}2.528 \\
256 &\cellcolor{red!20}2.750 &\cellcolor{red!21}2.566 &\cellcolor{red!19}2.527 \\
\midrule 
\multicolumn{4}{c}{$\alpha=1/8$} \\
4 &2.727 &2.549 &2.515 \\
8 &\textbf{2.725} &\textbf{2.545} &\textbf{2.510} \\
16 &{2.729} &2.550 &2.516 \\
32 &2.732 &2.553 &2.519 \\

64 &2.735 &2.553 &2.519 \\
128 &\cellcolor{red!18}2.743 &\cellcolor{red!19}2.564 &\cellcolor{red!20}2.530 \\
256 &\cellcolor{red!23}2.748 &\cellcolor{red!22}2.567 &\cellcolor{red!21}2.531 \\\midrule
8192 &\cellcolor{red!55}2.780 &\cellcolor{red!45}2.590 &\cellcolor{red!39}2.549 \\
\bottomrule
\end{tabular}

\end{table*}

\paragraph{Compute Budget} We show that when the total number of tokens is limited, our method can still improve language model performance. In Table~\ref{tab:scaling-compute}, we choose 12.5B, 25B, and 50B total tokens as the computing budget, and vary the expansion rate so that $w$ reaches $L$ at the same point during training. We observe that under different token budgets, the performance trend is the best: gradually expanding the context window gives better performance than a rapid increase.

\begin{table*}[tb]\centering

%\scriptsize
\small
\caption{Final validation loss under different training token budgets and expansion rate $\alpha$ with 120M models. $L_e$ represents the context length used for evaluation. ``\% of Token Budget'' means how many tokens are spent in the expansion phase with $w(t)$ increasing. Under all token budgets, we observe a consistent improvement when we spend around 64\% in expansion, and 36\% in the stable phase. }
\label{tab:scaling-compute}
\begin{tabular}{cccccc}\toprule
$\alpha$ &Tokens to $L$ (B) &\% of Token Budget &$L_e=512$ &$L_e=4096$ &$L_e=8192$ \\\midrule
\multicolumn{6}{c}{\textit{Token Budget = 12.5B}} \\
1 &8 &64\% &\textbf{2.912} &\textbf{2.732} &\textbf{2.698} \\
2 &4 &32\% &2.933 &2.746 &2.709 \\
4 &2 &16\% &2.958 &2.767 &2.729 \\
8 &1 &8\% &2.976 &2.782 &2.743 \\
\multicolumn{3}{c}{Baseline} &3.008 &2.823 &2.790 \\
\midrule
\multicolumn{6}{c}{\textit{Token Budget = 25B}} \\
1/2 &16 &64\% &\textbf{2.829} &\textbf{2.650} &\textbf{2.617} \\
1 &8 &32\% &2.841 &2.656 &2.619 \\
2 &4 &16\% &2.851 &2.665 &2.626 \\
4 &2 &8\% &2.873 &2.683 &2.645 \\
\multicolumn{3}{c}{Baseline} &2.918 &2.734 &2.700 \\
\midrule
\multicolumn{6}{c}{\textit{Token Budget = 50B}} \\
1/4 &32 &64\% &\textbf{2.771} &\textbf{2.590} &\textbf{2.556} \\
1/2 &16 &32\% &2.781 &2.596 &2.560 \\
1 &8 &16\% &2.789 &2.603 &2.564 \\
2 &4 &8\% &2.795 &2.607 &2.567 \\
\multicolumn{3}{c}{Baseline} &2.839 &2.652 &2.616 \\
\bottomrule
\end{tabular}

\end{table*}

\paragraph{Sliding Window Expansion}
A possible alternative to \ourmethod (using local causal masks by default) is to use a sliding window attention with a window size of $w(t)$ that changes with the training time. Formally, the mask becomes:
\[
M_{i,j} = 
\begin{cases} 
0 & \text{if } i-w \leq j \leq i \\
% 0 & \text{if } 0 \leq i-j \leq i\bmod w \\
-\infty & \text{otherwise.}
\end{cases}
\]
so that each token in the context has a fixed preceding context of size $w$. When $w(t)$ reaches $L$, the mask becomes equivalent to a causal mask. We compare the performance of the two in Table~\ref{tab:sliding-window-ladder-perf} and observe that the sliding window approach shows slightly better performance in long tasks and worse performance in standard benchmarks. This is likely because overall there are more tokens with longer preceding contexts for the sliding window approach. In both cases, \ourmethod outperforms the Random baseline. We think that future work could further investigate the differences between \ourmethod implementations with causal and sliding window attention, such as the formation of attention sink~\citep{gu2024attentionsinkemergeslanguage}. There could also be possible combinations of the two: for instance, using a local mask first to disable distraction, and enabling sliding windows as the training progresses. 

% \begin{wraptable}{r}
% {0.5\linewidth}

\begin{table*}
\small
\centering
\caption{Performance (\%) of 1B models with different masking schemes. All models are trained on the same 100B FineWeb-Pro tokens with a final context length of 8K. Both implementations of \ourmethod outperform the baseline, and the sliding window approach excels at long tasks with a slight performance drop on standard benchmarks.}\label{tab:sliding-window-ladder-perf}
\begin{tabular}{lll}\toprule
Model &Standard Avg. &Long Avg. \\\midrule
Random &52.5 &11.1 \\
+~\ourmethod w/ local causal &\textbf{55.2} {\scriptsize (+2.7)} &{12.3 }{\scriptsize (+1.2)} \\
+~\ourmethod w/ sliding window &54.4 {\scriptsize (+1.9)} &\textbf{12.8} {\scriptsize (+1.7)} \\
\bottomrule
\end{tabular}

\end{table*}
% \end{wraptable}
% Another work that shares similar strategies for training acceleration is ~\citet{jin2023growlengthacceleratingllmspretraining}, but their method does not give performance improvement when the number of training tokens is fixed. Moreover, they only consider training loss as a metric, while we comprehensively evaluate models of larger sizes on downstream tasks. 

\subsection{Additional Comparison with Related Work}\label{app:sec:additional-compared-with-related-work}

We acknowledge that there are several prior works discovering a similar pattern of short-to-long pretraining. For instance, ~\citet{li2022stabilityefficiencydilemmainvestigatingsequence} discover that using a sequence-length warmup for the initial steps in pretraining improves model stability. However, they mostly focus on stability in training loss and do not show a clear performance gain across multiple evaluations and larger scales. Moreover, we demonstrate that the benefits of scheduling a model's context window go beyond only the warmup stage. In Table~\ref{tab:expansion-rate-ablation}'s first row, simply warming up the model with 8B tokens results in suboptimal performance compared to a slower expansion rate. This validates that the context window should be considered as a factor to schedule over the entire training course, which also differentiates us from ~\citet{li2022stabilityefficiencydilemmainvestigatingsequence} that only consider the warmup stage. 

Another related work is \citet{jin2023growlengthacceleratingllmspretraining} where the authors use progressive sequence lengths to accelerate training. However, their method leads to worse performance under the same token budget, while our \ourmethod shows both time saving and performance improvement with the same number of tokens. We suspect that this might be because of the suboptimal schedule they used. 
Moreover, their study is limited to observing the training loss of small models (up to 410M parameters), while we comprehensively show performance gain across multiple corpora, model sizes, context sizes, and a wide variety of tasks. Overall, we systematically conduct controlled experiments on the impact of context window scheduling in pretraining, providing insights to explain these previous studies. 

\subsection{Compute Information}
\label{app:sec:compute-info}
We conducted all of our experiments for models with $\leq$ 1B size on an internal cluster of NVIDIA A100 nodes with 40G memory. Experiments with 3B models were conducted on H100 nodes.
There are additional preliminary experiments that we did not include in the paper, which account for a fraction of the total compute. The detailed computation for each experiment is as follows: 
For the preliminary study on context window, pretraining a 1B model with 100B tokens (with 8K context) takes around 200 hours on a node of 8 A100s. Models of different sizes scale accordingly. For instance, plotting Figure~\ref{fig:packing-length-compare}(a) and (b) requires a total of 159 days of pretraining on a single node. For \ourmethod experiments, the baseline pretraining using various corpora takes the same time, and SkyLadder speeds up the training by 13\% to 22\% depending on the context length.

\subsection{Dataset Statistics}

In this section, we provide detailed statistics of the datasets used in our study. These include the document length distributions of the pretraining corpora, the characteristics of the evaluation datasets, and the input length statistics of standard reasoning benchmarks. 

Table~\ref{tab:pretrain-dist} reports the document length statistics for the two pretraining corpora, \textit{CommonCrawl} and \textit{FineWeb-Pro}. Both distributions are strongly right-skewed, indicating that long documents are rare. Compared to FineWeb-Pro, CommonCrawl generally contains longer documents, while FineWeb-Pro has been more carefully cleaned and filtered.

\begin{table*}[h!]
\centering
\small
\caption{Document length statistics of the pretraining corpora, measured in tokens per document. 
Mean, median, and standard deviation describe the central tendency and variation. 
P25 and P75 indicate the 25th and 75th percentiles, while skewness and kurtosis capture distribution asymmetry and tail heaviness.}
\begin{tabular}{lrrrrrrrrr}
\toprule
Dataset & Mean & Median & StdDev & Min & Max & P25 & P75 & Skewness & Kurtosis \\
\midrule
CommonCrawl & 1973 & 1067 & 4567 & 45 & 594{,}272 & 651 & 1867 & 21 & 820 \\
FineWeb-Pro & 1364 & 849 & 2295 & 1 & 230{,}949 & 507 & 1481 & 15 & 533 \\
\bottomrule
\end{tabular}
\label{tab:pretrain-dist}
\end{table*}

Table~\ref{tab:benchmark-length} shows the input length characteristics of common reasoning and knowledge benchmarks, including ARC, CSQA, HellaSwag, OBQA, PIQA, SocialIQA, Winogrande, and MMLU. While these benchmarks consist of relatively short contexts, they remain standard for assessing a model’s factual consistency and reasoning ability. Importantly, a long-context model should maintain stable behavior even when the user provides a short query.

\begin{table*}[h!]
\centering
\caption{Input length characteristics of common reasoning and knowledge benchmarks. Although relatively short, these tasks are crucial for measuring knowledge and reasoning consistency.}
\small
\begin{tabular}{lrrrrrrrrr}
\toprule
Metric & ARC-C & ARC-E & CSQA & HellaSwag & OBQA & PIQA & SIQA & WinoG. & MMLU \\
\midrule
Mean & 222 & 216 & 146 & 508 & 129 & 224 & 226 & 149 & 540 \\
Std & 20 & 16 & 7 & 32 & 9 & 29 & 6 & 4 & 512 \\
Min & 191 & 194 & 134 & 435 & 116 & 191 & 210 & 140 & 155 \\
Max & 401 & 336 & 203 & 576 & 192 & 440 & 262 & 170 & 3144 \\
\bottomrule
\end{tabular}

\label{tab:benchmark-length}
\end{table*}

Finally, Table~\ref{tab:eval-stats} summarizes the characteristics of the evaluation datasets used in the reading comprehension and long-context evaluation. These include QA benchmarks such as MDQA, RULER, SQuAD, HotpotQA, NQ, TriviaQA, and RACE. The datasets differ substantially in input length, reflecting the diversity of reasoning depth and context complexity.

\begin{table*}[h!]
\centering
\small
\caption{Length statistics of reading comprehension and QA evaluation datasets. These benchmarks capture varying levels of input complexity, from short factual QA to multi-hop reasoning tasks.}
\begin{tabular}{lrrrrrrr}
\toprule
Metric & MDQA & RULER & SQuAD & HotpotQA & NQ & TriviaQA & RACE \\
\midrule
Mean & 5150 & 7259 & 1048 & 5010 & 583 & 566 & 492 \\
Std & 287 & 745 & 81 & 993 & 21 & 28 & 121 \\
Min & 4172 & 6209 & 923 & 3587 & 536 & 529 & 122 \\
Max & 6755 & 8061 & 1174 & 7842 & 633 & 643 & 1323 \\
\bottomrule
\end{tabular}

\label{tab:eval-stats}
\end{table*}

Overall, the datasets used in this work span a wide range of input lengths and domains, from large-scale pretraining corpora to short and long-context evaluation benchmarks, ensuring that our analysis is both comprehensive and representative.

\subsection{Licenses of Assets}\label{app:sec:license}
We mainly use the following public datasets or codebases in this paper: SlimPajama~\citep{cerebras2023slimpajama} following the CommonCrawl Foundation Terms of Use\footnote{https://commoncrawl.org/terms-of-use}, FineWeb-Pro~\citep{zhou2024programming} with an ODC-By 1.0 license, and TinyLlama~\citep{zhang2024tinyllamaopensourcesmalllanguage} with an Apache 2.0 License.

\end{document}